\documentclass[letterpaper]{article} 
\usepackage{aaai25}  
\usepackage{times}  
\usepackage{helvet}  
\usepackage{courier}  
\usepackage[hyphens]{url}  
\usepackage{graphicx} 
\usepackage{natbib}  
\usepackage{caption} 
\usepackage{subcaption}

\usepackage{amsmath}
\usepackage{algorithm}
\usepackage{algorithmicx}
\usepackage{algpseudocode}
\usepackage{multirow}
\usepackage{mathpazo}
\usepackage{xcolor}
\usepackage{endnotes}

\usepackage{newfloat}
\usepackage{listings}
\DeclareCaptionStyle{ruled}{labelfont=normalfont,labelsep=colon,strut=off} 
\floatstyle{ruled}
\newfloat{listing}{tb}{lst}{}
\floatname{listing}{Listing}
\title{TCAQ-DM: Timestep-Channel Adaptive Quantization for Diffusion Models}
\author{
    Haocheng Huang\textsuperscript{\rm 1,\rm 2,}\footnotemark[1], 
    Jiaxin Chen\textsuperscript{\rm 1,\rm 2,}\thanks{Equal Contribution. $^{\dag}$Corresponding Author.},
    Jinyang Guo\textsuperscript{\rm 3},
    Ruiyi Zhan\textsuperscript{\rm 1,\rm 2},
    Yunhong Wang\textsuperscript{\rm 1,\rm 2,$\dag$}
}
\affiliations {
    \textsuperscript{\rm 1}State Key Laboratory of Virtual Reality Technology and Systems, Beihang University, Beijing, China\\
    \textsuperscript{\rm 2}School of Computer Science and Engineering, Beihang University, Beijing, China\\
     \textsuperscript{\rm 3}School of Artificial Intelligence, Beihang University, Beijing, China
}

\usepackage{bibentry}

\begin{document}

\maketitle

\begin{abstract}
Diffusion models have achieved remarkable success in the image and video generation tasks. Nevertheless, they often require a large amount of memory and time overhead during inference, due to the complex network architecture and considerable number of timesteps for iterative diffusion. Recently, the post-training quantization (PTQ) technique has proved a promising way to reduce the inference cost by quantizing the float-point operations to low-bit ones. However, most of them fail to tackle with the large variations in the distribution of activations across distinct channels and timesteps, as well as the inconsistent of input between quantization and inference on diffusion models, thus leaving much room for improvement. To address the above issues, we propose a novel method dubbed Timestep-Channel Adaptive Quantization for Diffusion Models (TCAQ-DM). Specifically, we develop a timestep-channel joint reparameterization (TCR) module to balance the activation range along both the timesteps and channels, facilitating the successive reconstruction procedure. Subsequently, we employ a dynamically adaptive quantization (DAQ) module that mitigate the quantization error by selecting an optimal quantizer for each post-Softmax layers according to their specific types of distributions. Moreover, we present a progressively aligned reconstruction (PAR) strategy to mitigate the bias caused by the input mismatch. Extensive experiments on various benchmarks and distinct diffusion models demonstrate that the proposed method substantially outperforms the state-of-the-art approaches in most cases, especially yielding comparable FID metrics to the full precision model  on CIFAR-10 in the W6A6 setting, while enabling generating available images in the W4A4 settings.
\end{abstract}

\begin{figure}[!t]
    \centering
    \begin{subfigure}{.48\columnwidth}
        \centering
        \includegraphics[width=\linewidth]{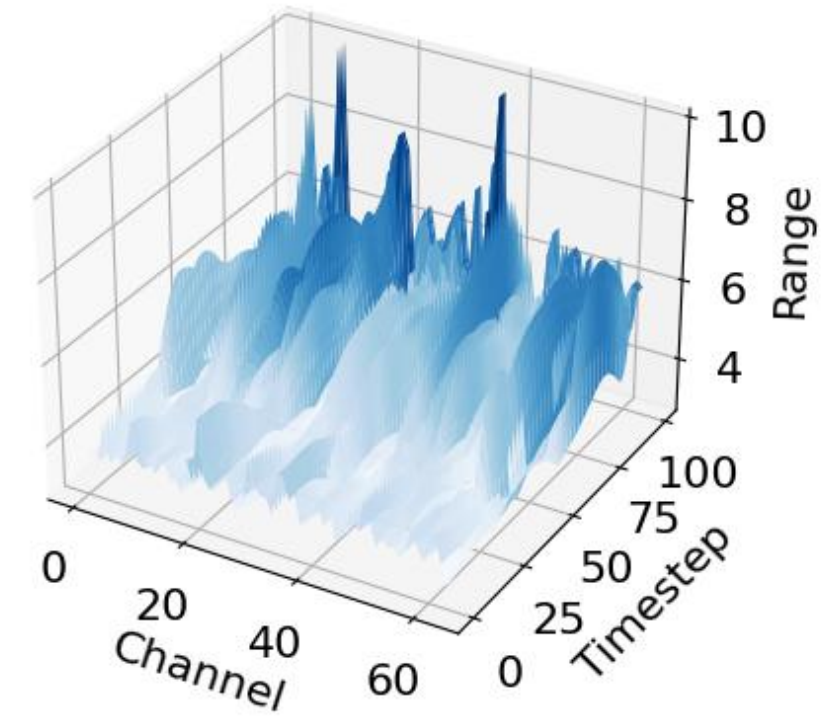}
        \caption{}
        \label{fig:problem_a}
    \end{subfigure}
    \begin{subfigure}{.48\columnwidth}
        \centering
        \includegraphics[width=\linewidth]{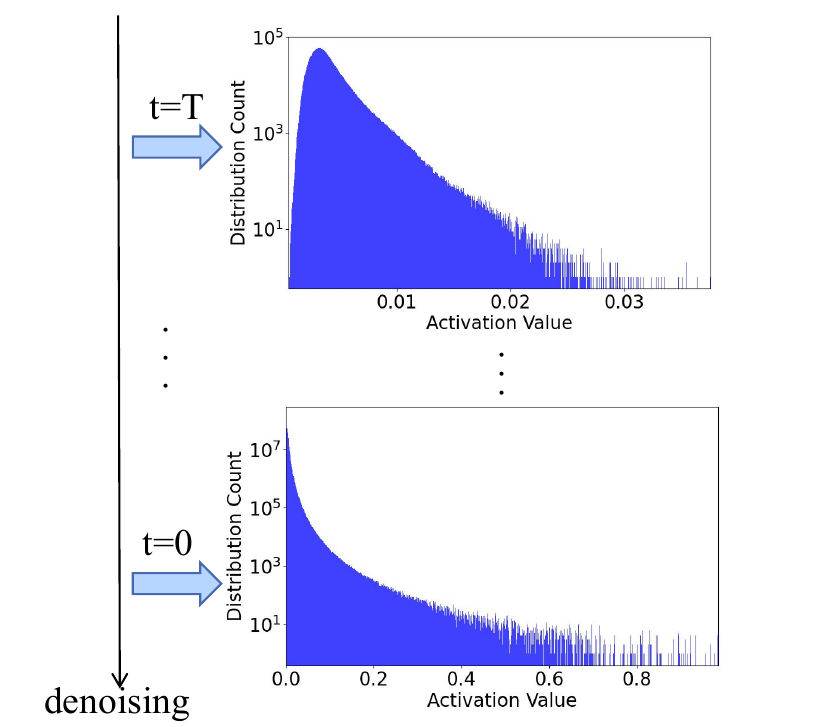}
        \caption{}
        \label{fig:problem_b}
    \end{subfigure}
    \begin{subfigure}{.96\columnwidth}
        \centering
        \includegraphics[width=\linewidth]{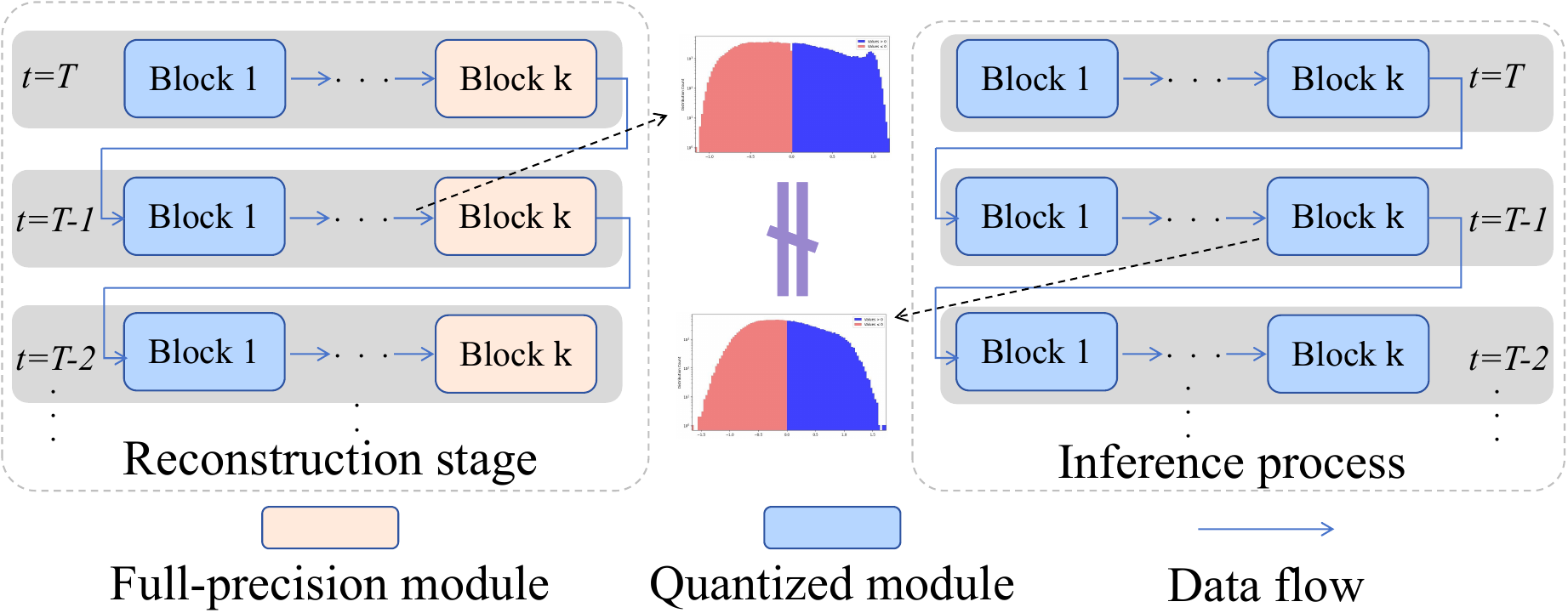}
        \caption{}
        \label{fig:problem_c}
    \end{subfigure}
    \caption{(a) Fluctuated activations per channels and timesteps in the convolutional layers (e.g. \textit{up.0.block.0.conv1} of DDIM). (b) Dynamic changes of activation distributions in the post-Softmax layer (e.g. \textit{down.1.attn.0} of DDIM) in distinct timesteps. (c) Misalignment between the intermediate data of the reconstruction stage in the quantization process and those in the inference process.}
    \label{fig:intro_problem}
\end{figure}

\section{Introduction}
Diffusion models \cite{DDPM} have emerged as one of the most prevailing generative models, with a wide range of applications including image generation \cite{DDPM,DDIM}, image translation \cite{dualdiffusion,plug}, super-resolution \cite{li2022srdiff,gao2023implicit,wang2024sinsr}, and video generation \cite{ho2022video,chen2024gentron}. They gradually transform noises into high-quality images or video clips through an iterative diffusion process, based on a noise estimation network and a denoising sampler. Nevertheless, due to the complex network structure and the massive network forward propagation required during dozens or even hundreds of iterative timesteps, existing models are generally computationally expensive, making it inefficient during inference.

Many efforts have been made to accelerate the diffusion model, which can be roughly divided into two categories. The first category of methods \cite{chung2022come,earlystop,howmuchisenough} focuses on decreasing the number of sampling timesteps. It is capable of linearly reducing the inference time cost without modifying the network structure, which however fails to decrease the model size. Alternatively, the second category of methods aims to expedite the inference through compressing the neural network by pruning \cite{ldpruner,laptop} and quantization \cite{PTQ4DM,APQ-DM,TFMQ-DM}. 

In this paper, we mainly investigate the post-training quantization (PTQ) technique for diffusion models, considering that it reduces both the storage and inference time cost by mapping the float-point weights and activations of the networks into low-bit integers \cite{BRECQ,adaround,fp8,dettmers2024qlora}, and is feasible for fast deployment without expensive re-training. Several works have attempted to explore the PTQ technique for diffusion models by either collecting calibration datasets across all timesteps \cite{PTQ4DM,Qdiff,TFMQ-DM} or correcting the accumulation errors on iterative sampling\cite{PTQD,TAC}. Nevertheless, most of them suffer from substantial performance degradation especially when quantizing under low bit-widths, as they fails to take the following characteristics of diffusion models into consideration based on our empirical observations: \textbf{1)} the range of activation in the convolutions layers often drastically fluctuates in both channels and timesteps as displayed in Fig.~\ref{fig:intro_problem}(a), which is prone to incur large quantization errors; \textbf{2)} the distribution of activations in the post-Softmax layers dynamically changes as the timestep decreases, and gradually exhibits a power-law-like shape during the diffusion process as shown in Fig.~\ref{fig:intro_problem}(b), resulting in nonnegligible quantization loss as most exiting works utilize one fix quantizer; \textbf{3)} existing reconstruction-based quantization methods often utilizes the output of the quantized model in the precedent block as input for reconstruction stage, which is not consistent with the inference process that adopt iterative sampling strategy as shown in Fig.~\ref{fig:intro_problem}(c), inevitably introducing bias and leaving much room for improvement.  

To address the above issues, we propose a novel PTQ approach dubbed Timestep-Channel Adaptive Quantization for Diffusion Models (TCAQ-DM). As displayed in Fig.~\ref{Fig:overview}, we first develop a timestep-channel joint reparameterization (TCR) module tailored for quantizing the convolutional layer with severely fluctuated activations. This module uniformly splits the overall timesteps into groups, and in each group balances the originally unconstrained activations by employing a channel-wise reparameterization transformation with timestep-aware average weighting. Subsequently, we present a dynamically adaptive quantizer (DAQ) specifically designed for quantizing the post-Softmax activations with timestep-varying distributions \cite{clauset2009power}. It establishes an estimator to assess the likelihood of the activations from a particular timestep obeying the power-law distribution on each layer. The timesteps with high likelihood are assigned a log2 quantizer \cite{li2023repq,fqvit}, which proved effective in quantizing the activations with power-law distributions, and those with low likelihood are dynamically handled by a uniform quantizer that is simple and efficient. Finally, to address the misalignment issue, we employ a progressively aligned reconstruction (PAR) strategy by incorporating the quantized inputs in the reconstruction stage of quantization process, in order to stay consistent with the inference process, thus further boosting the performance.


The main contributions of our work lie in three-fold:

\begin{itemize}
\item We propose a novel PTQ approach dubbed Timestep-Channel Adaptive Quantization for Diffusion Models (TCAQ-DM), by flexibly adapting to varying activation ranges and distributions in distinct channels and timesteps, and aligning the intermediate data in the quantization process with those in the inference process. 

\item We design a timestep-channel joint reparameterization (TCR) module to mitigate the influence of fluctuated activation ranges on quantization, and a dynamically adaptive quantizer (DAQ) to strengthen its flexibility in dealing with timestep-varying activation distributions in the post-Softmax layer, which reduces the quantization error especially under low bit-widths. We also develop a progressively aligned reconstruction (PAR) strategy to avoid the data inconsistency between quantization and inference, further boosting the performance.

\item We conduct extensive experiments and ablation studies on various datasets and representative diffusion models, and demonstrate that our method remarkably outperforms the state-of-the-art PTQ approaches for diffusion models in most cases, especially under low bit-widths. Particularly, for the challenging W4A4 setting, our method generate available results, while most compared PTQ approaches yield nearly collapsed performance. 

\end{itemize}

\section{Related Work}

Existing approaches for accelerating diffusion models roughly fall into two categories: building efficient diffusion models by reducing the sampling steps and compressing the network structures of diffusion models. For the later, we focus on the quantization based methods, and summarize the related works as below.

\subsection{Efficient Diffusion Model}
Diffusion models gradually apply Gaussian noise to real data in an iteratively process, as the preliminaries are provided in \cite{DDIM, PTQD}. For this process is time-consuming, many approaches have been proposed to obtain an efficient diffusion model by diminishing the sampling steps, which can be further divided into the training-based methods and the training-free ones. The former reduces the steps by model distillation \cite{knowledgediss,progressivedistillation,huang2024knowledge} or sample trajectory learning \cite{BDDM,learningfast,zhao2024unipc}. And the later usually directly designs efficient samplers on pre-trained diffusion models, by developing implicit samplers \cite{DDIM}, customized SDE, ODE solvers \cite{kim2023denoisingmcmc,zhang2022fast,zhou2024fast}, or automatic search \cite{AutoDiffusion}. Some methods also develop the cache-based strategies \cite{deepcache}. Despite decreasing the time cost, these methods fail to reduce the diffusion model size, thus still suffering from the high computational complexity and extensive storage consumption. 

\subsection{Model Quantization}
Different from the above approaches by reducing sampling steps, the model quantization alternatively aims at compressing the diffusion neural networks by mapping the float-point weights or activations into low-bit ones, thus decreasing both the inference latency and memory overhead. We review existing model quantization methods as below.

\subsubsection{General Quantization Methods}
Current quantization methods for general purpose mainly consists of the quantization-aware training (QAT) \cite{differentiable,root,chu2024make} and the post-training quantization (PTQ) \cite{adaround,BRECQ,qdrop,adalog}. The QAT methods mimic the quantization process and aim to reduce the quantization error during training. They often achieve high accuracy in low bit-width, but take tremendous training cost as they require retraining the overall weights on the whole large-scale training data. In contrast, the PTQ methods directly quantize the weight and activation based on a small-scale calibration set without finetuning the weights during quantization, thus being much more efficient due to its fewer data and computational costs. 

\subsubsection{PTQ for Diffusion Models}
Directly applying the general quantization methods to diffusion models usually results in poor performance. To deal with this problem, PTQ4DM \cite{PTQ4DM} collects calibration data from various timesteps, and makes the first attempt on quantizing diffusion models in 8 bit-width with slight performance degradation. Q-Diffusion \cite{Qdiff} further enhances the performance by dividing the skip connection layer. PTQD \cite{PTQD} eliminates the accumulation errors by correcting samplers and collecting the output at each timestep for calibration. APQ-DM \cite{APQ-DM} designs a dynamic grouping strategy and chooses a calibration set according to the structural risk minimization principle. TFMQ-DM \cite{TFMQ-DM} mitigates the information bias at different timesteps caused by quantization. PCR~\cite{ECCV24DM} proposes a progressive quantization method and an activation relaxing strategy, and TMPQ-DM \cite{TMPQ-DM} simultaneously reduces timesteps and quantize models. However, these methods fail to jointly handle the fluctuated activation ranges and distributions in distinct timesteps and channels, and neglect the inconsistency between the inputs of the reconstruction stage in the quantization process and those in the inference process, thus inclining to incur large quantization errors.

\begin{figure*}[!t]
    \centering
    \begin{subfigure}{2\columnwidth}
        \centering
        \includegraphics[width=\linewidth]{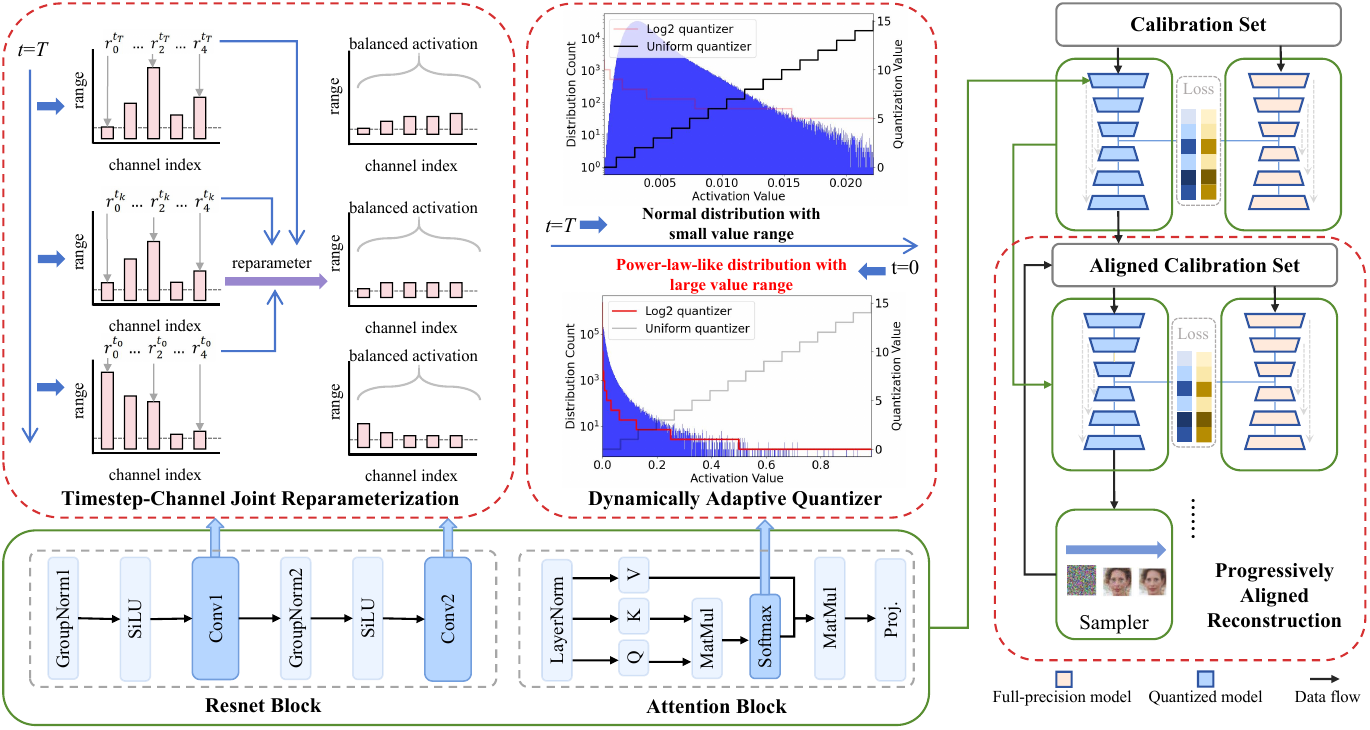}
        \label{fig:overview_sub1}
    \end{subfigure}

    \caption{Overview of our proposed method. In the initialization stage, we develop the timestep-channel joint reparameterization (TCR) and the dynamically adaptive quantizer (DAQ) to mitigate the fluctuated activation ranges in the convolutional layers, and the timestep-varying activation distributions in the post-Softmax layers, respectively. In the reconstruction stage, we design the progressively aligned reconstruction (PAR) strategy to further improve the generation performance by aligning the data flow in the quantization process with that in the inference process. }
    \label{Fig:overview}
\end{figure*}

\section{Methodology}
\subsection{Framework Overview}
Our method is based on PTQ that aims to compute the scaling factor $s$ and the zero point $z$, and map the float-point data to integers via the following formula:
\begin{equation}
  \mathbf{\hat{x}} = \Phi\left( \lfloor \frac{\mathbf{x}}{s} \rceil + z, 0, 2^{bit} - 1\right),  
\end{equation}
\noindent where $\mathbf{\hat{x}}$ denotes the quantized value of the float-point weights or activations $\mathbf{x}$, $\Phi$ indicates the function that clips the value range to $[0, 2^{bit} - 1]$, $\lfloor \cdot \rceil$ denotes the rounding operation, and $bit$ is the bit-width. Generally, by following existing works \cite{Qdiff,TFMQ-DM}, the overall quantization process includes the initialization stage that roughly search the quantization parameters, and the reconstruction stage that further refine the quantization parameters.

As shown in Fig.~\ref{Fig:overview}, different from current PTQ approaches, we propose three novel components, including the timestep-channel joint reparameterization (TCR) and the dynamically adaptive quantizer (DAQ) for the initialization stage, as well as the progressively aligned reconstruction (PAR) strategy for the reconstruction stage. TCR simultaneously mitigates the fluctuations of activation ranges in both distinct timesteps and channels in the convolutional layers, and DAQ adapts to varying activation distribution in the post-Softmax layers, both of which facilitate reducing the quantization error. PAR further boosts the performance by generating a calibration set that is aligned with the data flow in the inference process. The technical details are described in the rest part of this section.
\subsection{Timestep-Channel Joint Reparameterization} 
The activations of diffusion models' convolution layers exhibit significant fluctuation along the both timesteps and channels. The interplay between these two dimensions renders activation quantization considerably more arduous. To address this issue, we firstly group the activation quantization parameters uniformly under the inference timesteps:
\begin{align}
    S = \{s_0, s_1, ... s_{T-1} \},~~~Z = \{z_0, z_1, ...z_{T-1} \},
\end{align}
where $T$ denotes the denoising step, $S$ and $Z$ represent the scaling factor and zero point of the activation quantizer, respectively. It is noteworthy that the time cost of searching $S$ and $Z$ can be substantially diminished when leveraging a less-step sampling technique.

In addition to the timestep dimension, post-training quantization (PTQ) methods for diffusion models also suffer from activation variability across different channels. Inspired by recent quantization works in Vision Transformers (ViT) \cite{li2023repq}, we propose timestep-channel joint reparameterization module to solve this problem. This module rescales the input range by aggregating the values across all timesteps. Specifically, for a particular convolution layer with weights $W$ and input $\textbf{X}^t \in \mathbb{R}^{N \times C \times W \times H}$ of timestep $t$, we aim to find a scaling vector $\textbf{r}^t \in \mathbb{R}^C$, then reparameterize the activation and corresponding weight as:
\begin{equation}
\mathbf{X}_{:,j}^{t'} = \mathbf{X}_{:,j}^t \oslash \mathbf{r}^t_j,~~~
\mathbf{W}_{:,j}^{'} = \mathbf{W}_{:,j} \odot \mathbf{r}^t_j,
\end{equation}
where $\oslash$ and $\odot$ denote the broadcast division and multiplication, respectively. For a convolution layer, which is equivalent to a linear affine operation, this reparameterization will retain the output while shifting the value range from activations to weights. A tailored $\textbf{r}^t$ for activation $\textbf{X}^t$ aligns activations between channels:
\begin{equation}
\mathbf{r}^t_j = \max(\mathbf{X}_{:,j})/s^t_{tar} , 
\end{equation}
where $s^t_{tar}$ is a pre-specific target range of timestep $t$. However, diffusion models have different activations between timesteps while sharing the same weights. Rescaling for each timestep will produce multiple weights, causing a high storage cost. Therefore, it is necessary to combine $\mathbf{r}^t$ of all timesteps to general scaling vector $\mathbf{r}$.


When performing reparameterization, we find some of the channels have a small value range in most of the denoising steps, but suddenly increase in a few steps and become outliers. To limit these channels' value range, we first set the minimum of the maximum value of all channels as the rescale target $s^t$, to ensure that the value range of all channels will not be further expanded. Then, we use the maximum value of each channel as the weight to sum the activation across all timesteps, ensuring the scaling vector on the timestep with larger activation receives more attention. 
The final formula is shown as:
\begin{equation}
\begin{array}{c}
s_{tar}^{t} = \min(\max(\mathbf{X}^t_{:,d})_{1\leq d \leq D}),\\
\mathbf{r}_d^t = \frac{\max(\mathbf{X}^t_{:,d})}{s_{tar}^{t}},
\mathbf{r}_d^s = \frac{\sum_t r_d^t * \max(\mathbf{X}^t_{:,d})}{\sum_t \max(\mathbf{X}^t_{:,d})}, \\
\widetilde{\mathbf{X}}_{:,d}^{t} = \mathbf{X}_{:,d}^t \oslash \mathbf{r}^s_d, ~~~
\widetilde{\mathbf{W}}_{:,d} = \mathbf{W}_{:,d} \odot \mathbf{r}^s_d.
\end{array}
\end{equation}

Since this method will enlarge the weight values, which may lead to insignificant performance improvement when applying weight relative low-bit quantization like W4A8, we use a hyper-parameter $R_{tru}$ to truncate the scaling vector to a limit range in these settings.

\subsection{Dynamically Adaptive Quantizer}
The activation of post-Softmax often shows a power-law distribution. The uniform quantizer cannot balance the quantization between the long-tail and the small value peak of this type of distribution, which often leads to performance degradation. Previous methods \cite{fqvit} attempt to use a log2 quantizer to fit the feature of the post-Softmax. It maps the float numbers to a logarithmic function with a base of 2:
\begin{align}
    \mathbf{\hat{x}} = \Phi (\lfloor -\log_2 \frac{\mathbf{x}}{s} \rceil, 0, 2^{bit - 1}), 
    \mathbf{\tilde{x}} = s * 2^{-\mathbf{\hat{x}} },
\end{align}
where $\mathbf{\hat{x}}$ and $\mathbf{\tilde{x}}$ indicates quantized value and dequantized value of $\mathbf{x}$, respectively.


However, the post-Softmax activation in diffusion models also suffers from timestep variance. In the early denoising steps of certain blocks, activations is only distributed within a limited range, where the log quantizer will lead to a larger quantization error. Directly applying the log2 quantizer in diffusion models may even perform poorly in high-bit setting, as shown in Table \ref{tab:dynamic effect}. Therefore, we propose a dynamically adaptive quantizer that could select whether to use the log2 quantizer for a specific timestep of a post-Softmax layer, based on its mathematical properties. Specifically, we use the Maximum Likelihood Estimation method to fit the each layers activation on every timestep to a power-law distribution \cite{clauset2009power}:
\begin{align}
P(X \geq x) = cx^{-\alpha}.
\end{align}

Since the activations of model could be collected in advance, this operation could be conducted offline. Then we calculate the ratio of the likelihood estimation results for the power-law distribution and other distributions (e.g., log-normal or exponential) as $R_g$, and perform log2 quantizer to the specific timestep where the ratio is greater than zero:
\begin{equation}
    \mathbf{\hat{x}} = \begin{cases}
    \lfloor \frac{\mathbf{x}}{s_g} + z_g \rceil, & if ~R_g \leq 0; \\
    \lfloor -\log_2\frac{\mathbf{x}}{s_g} \rceil, & if ~R_g > 0.
    \end{cases}
\end{equation}

It is worth noting that DAQ performs offline, and only introduces a small amount of extra computational overhead, roughly 3\% of the overall cost, which is affordable in our implementation. 

\subsection{Progressively Aligned Reconstruction}

Existing PTQ methods often comprises a block reconstruction stage to improve the performance. However, the iterative inference process of diffusion models causes an inconsistency problem between the reconstruction stage and inference process, as shown in Fig. \ref{fig:problem_c}. Existing PTQ methods would introduce biased input distributions when applied to diffusion models, as the model is quantized in a single forward round during reconstruction, and is then called iteratively during the inference phase.

For diffusion models sharing weights across all timesteps, quantizing blocks in the same order as the denoising process is challenging. As an alternative method, we propose progressively aligned reconstruction to iteratively align the inputs. In particular, after the basic reconstruction with BRECQ \cite{BRECQ}, we continuously sample a new calibration set using the quantized model and then utilize this aligned set to reconstruct the model. This phase will repeated in multiple rounds with fewer iterations than the first one. We refer to the \emph{Supplementary Material} for the detailed algorithm of the proposed PAR method. \footnote{Supplementary Material is available at: \url{https://dr-jiaxin-chen.github.io/page/}}

\begin{table}[!t]
    \centering
    \begin{tabular}{c|c|c|c}
    \hline 
    Method & Bit-width & FID ($\downarrow$) & IS ($\uparrow$) \\
    \hline
    FP model & W32A32 & 4.14 & 9.12 \\
    \hline
    PTQ4DM & W4A32 & 5.65 & 9.02 \\
    Q-Diffusion & W4A32 & 5.09 & 8.78 \\
    TFMQ-DM & W4A32 & 4.73 & \textbf{9.14} \\
    \textbf{Ours} & W4A32 & \textbf{4.28} & 9.09 \\
    \hline
    PTQ4DM & W8A8 & 5.69 & 9.31 \\
    Q-Diffusion* & W8A8 & 4.78 & 8.89 \\
    APQ-DM & W8A8 & 4.24 & 9.07 \\
    TFMQ-DM & W8A8 & 4.24 & 9.07 \\
    TAC-Diffusion & W8A8 & \textbf{3.68} & \textbf{9.49} \\
    \textbf{Ours} & W8A8 & 4.09 & 9.08 \\
    \hline
    PTQ4DM & W4A8 & 10.12 & \textbf{9.31} \\
    Q-Diffusion & W4A8 & 4.93 & 9.12 \\
    TFMQ-DM & W4A8 & 4.78 & 9.13 \\
    TAC-Diffusion & W4A8 & 4.89 & 9.15 \\
    \textbf{Ours} & W4A8 & \textbf{4.59} & 9.17 \\
    \hline
    PTQ4DM* & W6A6 & 61.83 & 7.10 \\
    Q-Diffusion* & W6A6 & 26.06 & 9.02 \\
    TFMQ-DM* & W6A6 & 9.59 & 8.84 \\
    \textbf{Ours} & W6A6 & \textbf{4.40} & \textbf{9.04} \\
    \hline
    PTQ4DM* & W4A4 & 375.12 & 0.45 \\
    Q-Diffusion* & W4A4 & 384.21 & 0.71 \\
    TFMQ-DM* & W4A4 & 236.63 & 3.19 \\
    \textbf{Ours} & W4A4 & \textbf{6.38} & \textbf{8.70} \\
    \hline
    \end{tabular}
    \caption{Comparison results on CIFAR-10 based on DDIM model with 100 timesteps. * means directly rerunning the open-resource code.}
    \label{tab: CIFAR-results}
\end{table}

\begin{figure}[!t]
    \centering
    \begin{subfigure}[t]{.15\columnwidth}
        \centering
        \includegraphics[width=\linewidth]{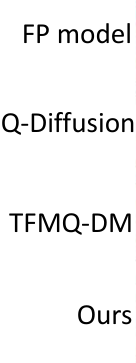}
        \label{fig:vis_title}
    \end{subfigure}
    \begin{subfigure}[t]{.4\columnwidth}
        \centering
        \includegraphics[width=\linewidth]{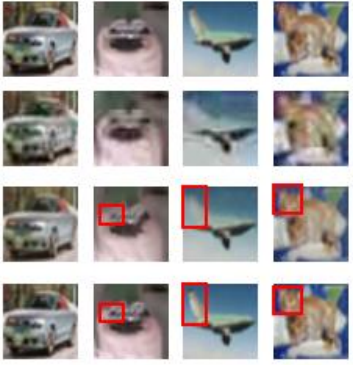}
        \caption{W6A6}
        \label{fig:vis_w6a6}
    \end{subfigure}
    \hfill
    \begin{subfigure}[t]{.4\columnwidth}
        \centering
        \includegraphics[width=1\linewidth]{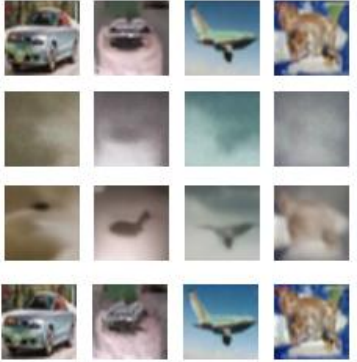}
        \caption{W4A4}
        \label{fig:vis_w4a4}
    \end{subfigure}
    \caption{Visualization of images generated by quantized models via various PTQ methods, indicating that our method generates images with better visual details in W6A6, and outputs available images in the challenging W4A4 setting.}
    \label{fig:vis}
\end{figure}



\section{Experimental Results and Analysis}

\subsection{Experimental Settings}
By following existing works \cite{Qdiff,TFMQ-DM}, we evaluate our proposed method  on the \textit{ImageNet} dataset \cite{imagenet} by using LDM-4 for the conditional generation task. For the unconditional generation task, we conduct experiments on the \textit{CIFAR-10} dataset \cite{cifar} by using DDIM \cite{DDIM}, \textit{LSUN-Bedrooms} and \textit{LSUN-Churches} dataset \cite{lsun} based on LDM-4. 
Similar to \cite{Qdiff,TFMQ-DM}, we adopt the evaluation metrics including Fréchet Inception Distance (FID) \cite{fid} and Inception Score (IS) \cite{is} on CIFAR-10 and ImageNet, and additionally report sliced FID (sFID) \cite{is} when using LDM. FID and sFID are based on calculating the mean and covariance of features extracted by the Inception network, while IS depends on the predicted distribution from the classification model.

\subsection{Implementation Details}
By following \cite{Qdiff,TFMQ-DM}, we perform the channel-wise quantization on weights and layer-wise quantization on activations. Similar to \cite{TFMQ-DM}, we maintain the input and output layers of the model in full precision and generate calibration sets by the full-precision model. For the weight quantization, we conduct BRECQ with 20,000 iterations for initialization, and 10,000 iterations for each progressive round with a batch size of 16. For the activation quantization, we use the commonly used hyper-parameter search method as depicted in RepQ-ViT \cite{li2023repq} with a batch size of 64. All experiments are conducted with an 8-bit post-Softmax layer unless being specifically claimed. 
All experiments are conducted on a single RTX4090 GPU.

\begin{table}[!t]
    \centering
    \begin{tabular}{c|c|c|c}
    \hline 
    Method & Bit-width  & FID ($\downarrow$) & sFID ($\downarrow$)\\
    \hline
    FP model & W32A32 S32 & 2.98 & 7.09 \\
    \hline
    Q-Diffusion & W4A32 S8 & 4.20 & 7.66 \\
    PTQD & W4A32 S8 & 4.42  & 7.88 \\
    TFMQ-DM & W4A32 S32 &3.60 & 7.61 \\
    \textbf{Ours} & W4A32 S8 & \textbf{3.55} & \textbf{7.54} \\
    \hline
    Q-Diffusion & W8A8 S8 & 4.51 & 8.17 \\
    PTQD & W8A8 S8 & 3.75 & 9.89 \\
    TFMQ-DM & W8A8 S8 & 3.14  & \textbf{7.26} \\
    \textbf{Ours} & W8A8 S8 & 3.21 & 7.59 \\
    \textbf{Ours} & W8A8 S32 & \textbf{3.11} & 7.34 \\
    \hline
    Q-Diffusion & W4A8 S8 & 6.40 & 17.93 \\
    PTQD & W4A8 S8 & 5.94 & 15.16 \\
    TFMQ-DM & W4A8 S32  & 3.68 & \textbf{7.65} \\
    TAC-Diffusion & W4A8 S8 & 4.94 & - \\
    \textbf{Ours} & W4A8 S8 & 3.70 & 7.69 \\
    \textbf{Ours} & W4A8 S32 & \textbf{3.65} & 7.67 \\
    \hline
    Q-Diffusion* & W4A4 S8 & 334.83 & 190.89 \\
    PTQD* & W4A4 S8 & 321.47 & 181.61 \\
    TFMQ-DM* & W4A4 S32 & 118.70 & 80.85 \\
    \textbf{Ours} & W4A4 S8 & \textbf{16.43} & \textbf{23.85} \\
    \hline
    \end{tabular}
    \caption{Comparison results on LSUN-bedrooms with LDM-4 using the DDIM sampler with 200 timesteps.}
    \label{tab:bedrooms-results}
\end{table}

\begin{table}[ht]
    \centering
    \begin{tabular}{c|c|c|c}
    \hline 
    Method & Bit-width & FID ($\downarrow$) & sFID ($\downarrow$) \\
    \hline
    FP model & W32A32 & 4.08 & 10.89 \\
    \hline
    Q-Diffusion & W4A32 & 4.55 & 11.90 \\
    PTQD & W4A32 & 4.67 & 13.68 \\
    TFMQ-DM & W4A32 & 4.07 & \textbf{11.41} \\
    \textbf{Ours} & W4A32 & \textbf{4.00} & 11.72 \\
    \hline
    Q-Diffusion & W8A8 & 4.87 & 12.23 \\
    PTQD & W8A8 & 4.89 & 12.23 \\
    TFMQ-DM & W8A8 & \textbf{4.01} & 10.98 \\
    \textbf{Ours} & W8A8 & 4.05 & \textbf{10.82} \\
    \hline
    Q-Diffusion & W4A8 & 4.66 & 13.97 \\
    PTQD & W4A8 & 5.10 & 13.23 \\
    TFMQ-DM & W4A8 & 4.14 & \textbf{11.46} \\
    \textbf{Ours} & W4A8 & \textbf{4.13} & 11.57 \\
    \hline
    Q-Diffusion* & W4A4 & 360.32 & 191.75 \\
    PTQD* & W4A4 & 358.34 & 180.26 \\
    TFMQ-DM* & W4A4 & 236.52 & 186.44 \\
    \textbf{Ours} & W4A4 & \textbf{29.17} & \textbf{35.89} \\
    \hline
    \end{tabular}
    \caption{Comparison results on LSUN-Churches based on LDM-4 by the DDIM sampler with 500 timesteps.}
    \label{tab:church-results}
\end{table}

\begin{table}[!t]
    \centering
    \small
    \begin{tabular}{c|c|c|c|c}
    \hline 
    Method & Bit-width & FID ($\downarrow$) & IS ($\uparrow$) & sFID ($\downarrow$) \\
    \hline
    FP model & W32A32 & 10.91 & 235.64 & 7.67 \\
    \hline
    Q-Diffusion & W4A32 & 11.87 & 213.56 & 8.76 \\
    PTQD & W4A32 & 11.65 & 210.78 & 9.06 \\
    TFMQ-DM & W4A32 & 10.50 & 223.81 & 7.98 \\
    \textbf{Ours} & W4A32 & \textbf{10.50} & \textbf{234.51} & \textbf{6.66} \\
    \hline
    Q-Diffusion & W8A8 & 12.80 & 187.65 & 9.87 \\
    PTQD & W8A8 & 11.94 & 153.92 & 8.03 \\
    TFMQ-DM & W8A8 & 10.79 & 198.86  & 7.65 \\
    \textbf{Ours} & W8A8 & \textbf{10.58} & \textbf{239.41} & \textbf{7.54} \\
    \hline
    Q-Diffusion & W4A8 & 10.68 & 212.51 & 14.85 \\
    PTQD & W4A8 & 10.40 & 214.73 & 12.63 \\
    TFMQ-DM & W4A8 & 10.29 & 221.82 & \textbf{7.35} \\
    \textbf{Ours} & W4A8 & \textbf{9.97} & \textbf{232.87} & 7.67 \\
    \hline
    Q-Diffusion* & W4A4 & 376.54 & 1.69 & 165.39 \\
    PTQD* & W4A4 & 361.29 & 1.87 & 190.48 \\
    TFMQ-DM* & W4A4 & 210.06 & 2.95 & 192.81 \\
    \textbf{Ours} & W4A4 & \textbf{30.69} & \textbf{86.11} & \textbf{18.92} \\
    \hline
    \end{tabular}
    \caption{Comparison results on ImageNet based on LDM-4 by using the DDIM sampler with 20 timesteps.}
    \label{tab:in-results}
\end{table}

\subsection{Comparison to the State-of-the-Art Methods}

 We compare our method to the state-of-the-art PTQ approaches, including PTQ4DM \cite{PTQ4DM}, Q-Diffusion \cite{Qdiff}, PTQD \cite{PTQD}, APQ-DM \cite{APQ-DM}, TFMQ-DM \cite{TFMQ-DM} and TAC-Diffusion \cite{TAC}.

\subsubsection{Unconditional Image Generation.}
On CIFAR-10 with DDIM, we follow the same setting as Q-Diffusion \cite{Qdiff}. As shown in Table \ref{tab: CIFAR-results} and Fig.~\ref{fig:vis}, our method reaches competitive FIDs compared to the full-precision model, and outperforms the state-of-the-art approaches in most cases. As for W4A4, the performance of existing approaches is collapsed with large FIDs. In contrast, our method achieves promising results with less than 2.3 increase in FID, compared to the full-precision model. We provide more visualization results in the \emph{Supplementary Material}.

On LSUN-bedrooms with Latent Diffusion Model (LDM), we adopt the same settings as TFMQ-DM \cite{TFMQ-DM}, except for the post-Softmax quantization bit-width. As shown in Table \ref{tab:bedrooms-results}, our method with 8-bit post-Softmax significantly promotes the FID, compared to the other methods using the same setting. And when using 32 bits, our method is comparable to TFMQ-DM in most cases, and remarkably outperforms it under the W4A4 setting.

\subsubsection{Conditional Image Generation.}

On ImageNet, we employ a denoising process with 20 iterations, following the same setting as TFMQ-DM. As shown in Table \ref{tab:in-results}, our method improves FIDs of TFMQ-DM by 0.21 and 1.32 under W8A8 and W4A32, respectively. As for the challenging W4A4 settings, despite that there is a gap between the full-precision model and the quantized model, our method still reaches a comparable performance, while the compared methods performs poorly with extremely large FIDs and sFIDs.

\begin{table}[!t]
    \centering
    \begin{tabular}{c|c|c|c}
    \hline 
    Method & Bit-width & FID ($\downarrow$) & IS ($\uparrow$) \\
    \hline
    Baseline & W8A8 & 4.78 & 8.87 \\
    +TCR & W8A8 & 4.12 & 9.04 \\
    +TCR+DAQ & W8A8 & 4.11 & 9.06 \\
    +TCR+DAQ+PAR & W8A8 & \textbf{4.09} & \textbf{9.08}  \\
    \hline
    Baseline & W4A8 & 4.93 & 9.12 \\
    +TCR & W4A8 & 4.76 & 9.02 \\
    +TCR+DAQ & W4A8 & 4.59 & 8.97 \\
    +TCR+DAQ+PAR & W4A8 & \textbf{4.59} & \textbf{9.17} \\
    \hline
    Baseline & W6A6 & 26.60 & 9.02 \\
    +TCR & W6A6 & 4.59 & 8.99 \\
    +TCR+DAQ & W6A6 & 4.47 & \textbf{9.09} \\
    +TCR+DAQ+PAR & W6A6 & \textbf{4.40} & 9.04 \\
    \hline
    Baseline & W4A4 & 371.61 & 0.41 \\
    +TCR & W4A4 & 9.54 & 8.57 \\
    +TCR+DAQ & W4A4 & 9.09 & 8.37 \\
    +TCR+DAQ+PAR & W4A4 & \textbf{6.38} & \textbf{8.70} \\
    \hline
    \end{tabular}
    \caption{Ablation results of the proposed main components on CIFAR-10 based on DDIM with 100 timesteps.}
    \label{tab:ablation}
\end{table}

\begin{table}[!t]
    \centering
    \begin{tabular}{c|c|c|c}
    \hline 
    Method & Bit-width & FID ($\downarrow$) & IS ($\uparrow$) \\
    \hline
    Log2 quantizer & W6A6 S8 & 4.97 & 8.95 \\
    Uniform quantizer & W6A6 S8 & 4.66 & \textbf{9.08} \\
    \textbf{DAQ (Ours)} & W6A6 S8 & \textbf{4.42} & 9.04 \\
    \hline
    Log2 quantizer & W6A6 S6 & 4.77 & \textbf 8.63 \\
    Uniform quantizer & W6A6 S6 & 4.87 & 8.53 \\
    \textbf{DAQ (Ours)} & W6A6 S6 & \textbf{4.61} & \textbf{9.05} \\
    \hline
    Log2 quantizer & W6A6 S4 & 4.76 & 9.01 \\
    uniform quantizer & W6A6 S4 & 14.20 & 8.06 \\
    \textbf{DAQ (Ours)} & W6A6 S4 & \textbf{4.66} & \textbf{9.07} \\
    \hline
    \end{tabular}
    \caption{Ablation results on distinct quantizers under different post-Softmax bit-widths on CIFAR-10 based on DDIM with 100 timesteps. }
    \label{tab:dynamic effect}
\end{table}

\subsection{Ablation Study}
To evaluate the effectiveness of each proposed component, we perform ablation study on CIFAR-10 based on DDIM, by employing BRECQ as the baseline method. 

\subsubsection{Effectiveness of TCR}
As summarized in Table \ref{tab:ablation}, the proposed TCR module reduces the FID by 0.66 and 22.01 in the W8A8 setting and the W6A6 setting respectively, compared to the baseline. Moreover, this module plays a crucial role in maintaining a comparable performance in low-bit quantization such as W4A4, with a substantial improvement of FID. 

\subsubsection{Effectiveness of DAQ}
In terms of the DAQ module, it further promotes the performance across all bit-widths, especially improving the FID by 0.17 and 0.45 in the W4A8 and W4A4 settings, respectively. Moreover, as displayed in Table \ref{tab:dynamic effect}, DAQ achieves stable improvements across all Softmax bit-widths, compared with the uniform and log2 quantizers.


\subsubsection{Effectiveness of PAR}
As shown in Table \ref{tab:ablation}, the proposed PAR module also obtains improvements, reducing the FID significantly by 2.71 in the W4A4 setting and promoting the performance in other bit-widths. 

More experimental results about hyper-parameters are provided in the \emph{Supplementary Material}.

\section{Conclusion}
In this work, we propose a novel post-training quantization method, dubbed Timestep-Channel Adaptive Quantization for Diffusion Models (TCAQ-DM). We first develop the timestep-channel joint reparameterization (TCR) to mitigate the fluctuated activation ranges. Subsequently, we employ a dynamically adaptive quantizer (DAQ) to reduce the quantization errors caused by the timestep-varying activation distributions. Moreover, we design a progressively aligned reconstruction (PAR) strategy to align the data in the reconstruction stage of the quantization process with that during inference, further boosting the performance. Extensive experimental results on distinct dataset and diffusion models as well as extensive ablation results clearly demonstrate the superiority of the proposed approach under low bit-widths.

\section{Acknowledgements}

This work was partly supported by the National Natural Science Foundation of China (Nos. 62202034, 62306025, 92367204), the Beijing Natural Science Foundation (No. 4242044), the Beijing Municipal Science and Technology Project (No. Z231100010323002), the Aeronautical Science Foundation of China (No. 2023Z071051002), the Research Program of State Key Laboratory of Virtual Reality Technology and Systems, and the Fundamental Research Funds for the Central Universities.

\end{document}


\maketitle

In this document, we provide more illustration on the fluctuation of activation in distinct timesteps and channels, additional visualization results on the effectiveness of the proposed TCR module, extra details and results of the effect of the proposed PAR module, as well as more quantitative results by various quantization methods on distinct datasets.

\subsection{Illustration on Activation Fluctuation}
\begin{figure*}[!t]
    \centering
    \begin{subfigure}{\columnwidth}
        \centering
        \includegraphics[width=0.85\linewidth]{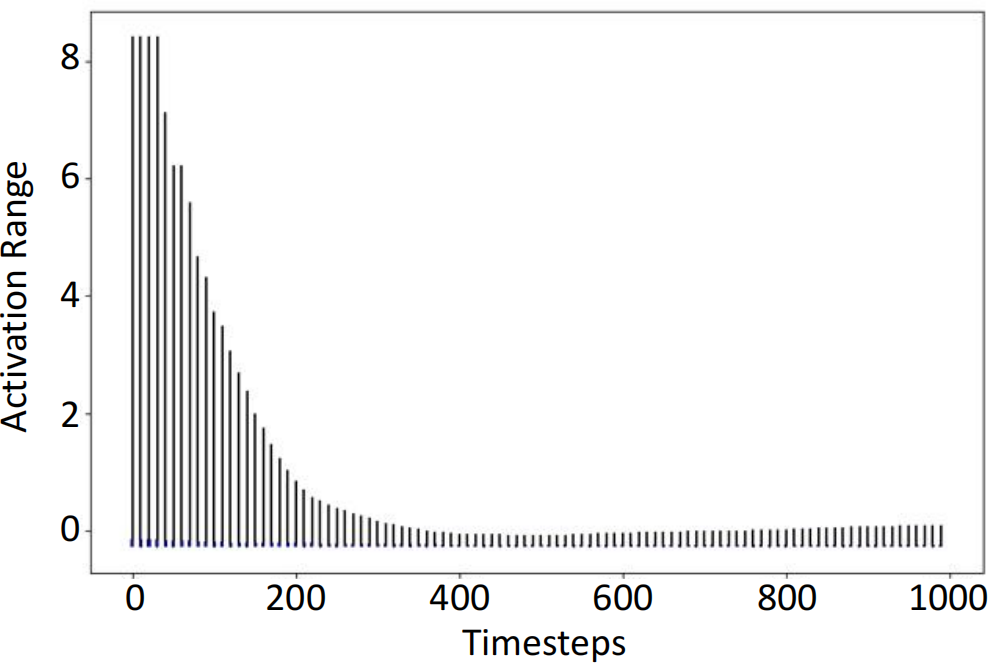}
        \caption{Activations at distinct timesteps from the 177th channel of \textbf{up.0.block.1.conv1}.}
        \label{fig:time_sub1}
    \end{subfigure}
    \begin{subfigure}{\columnwidth}
        \centering
        \includegraphics[width=0.85\linewidth]{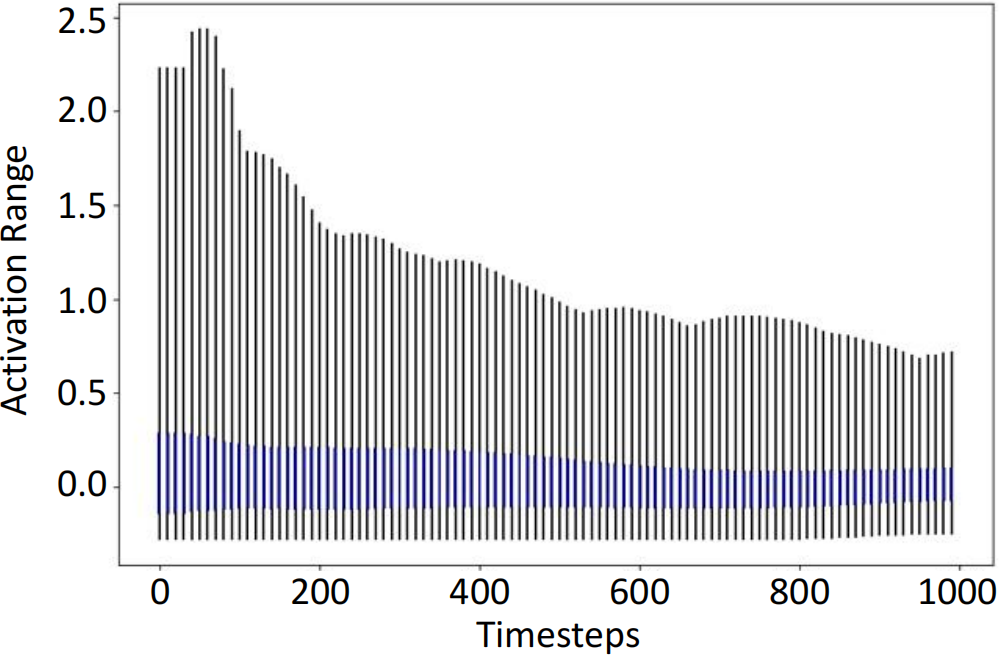}
        \caption{Activations at distinct timesteps from the 91st channel of \textbf{up.0.block.1.conv1}.}
        \label{fig:time_sub2}
    \end{subfigure}
    \caption{(a) The value of certain channels suddenly spikes during inference, making reparameterization more challenging. (b) Most channels are located within a narrow range.}
    \label{fig:time}
\end{figure*}

As discussed in the main body, the activations in certain channels exhibit a narrow value range throughout most of the denoising timesteps. However, they abruptly spike during a few steps, becoming outliers, as displayed in Fig. \ref{fig:time}. Since the averaged value remains confined to a limited range, these outliers are further amplified if the minimum value is not used as the rescale target. In contrast, if the average weight strategy is omitted, the values are no longer constrained to a narrow range, resulting in suboptimal performance.

\subsection{Visualization of TCR}
\begin{figure*}[!t]
    \centering
    \begin{subfigure}{2\columnwidth}
        \centering
        \begin{subfigure}{.3\columnwidth}
            \centering
            \includegraphics[width=\linewidth]{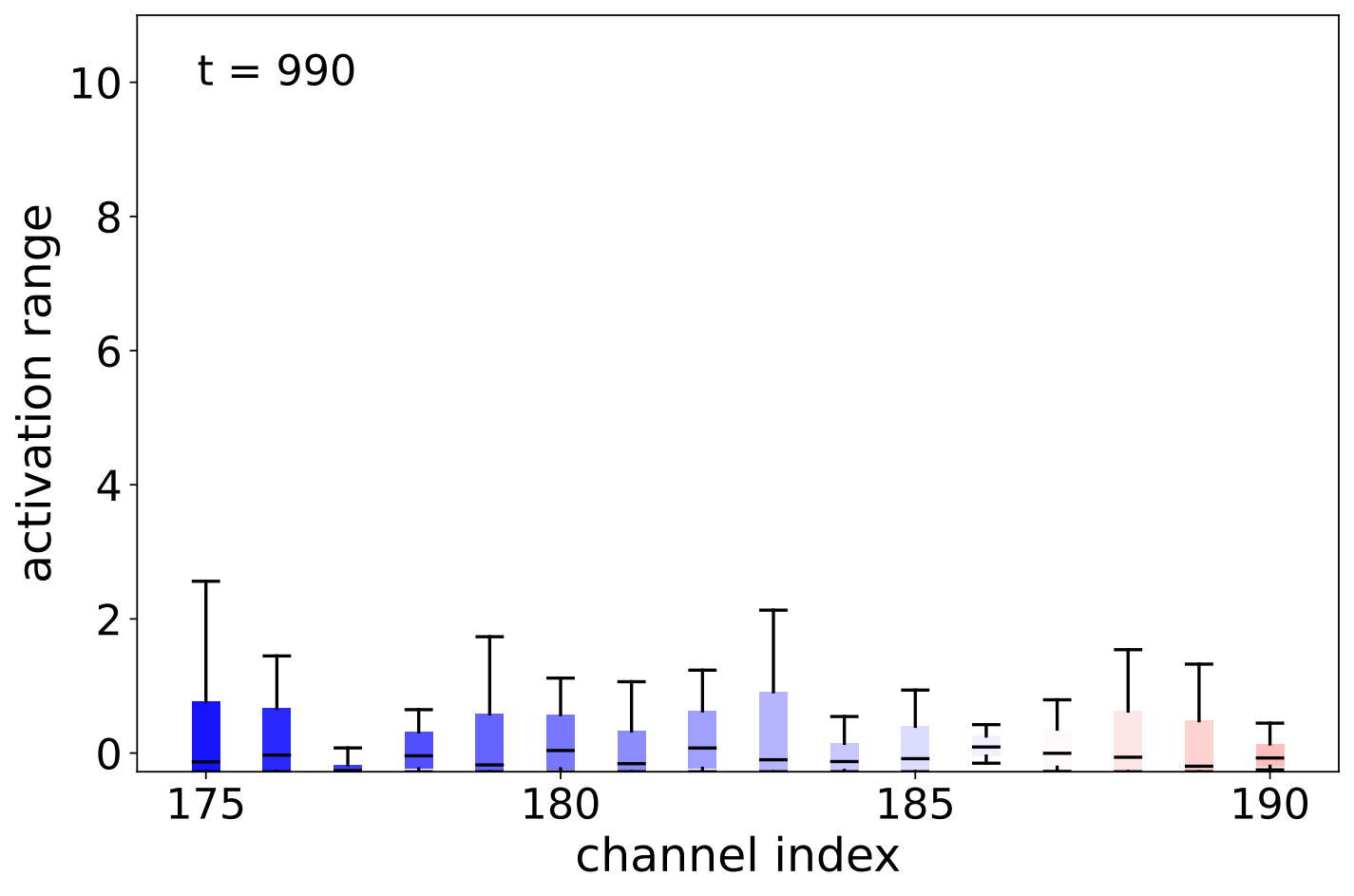}
            \label{fig:dist_sub1}
        \end{subfigure}
        \begin{subfigure}{.3\columnwidth}
            \centering
            \includegraphics[width=\linewidth]{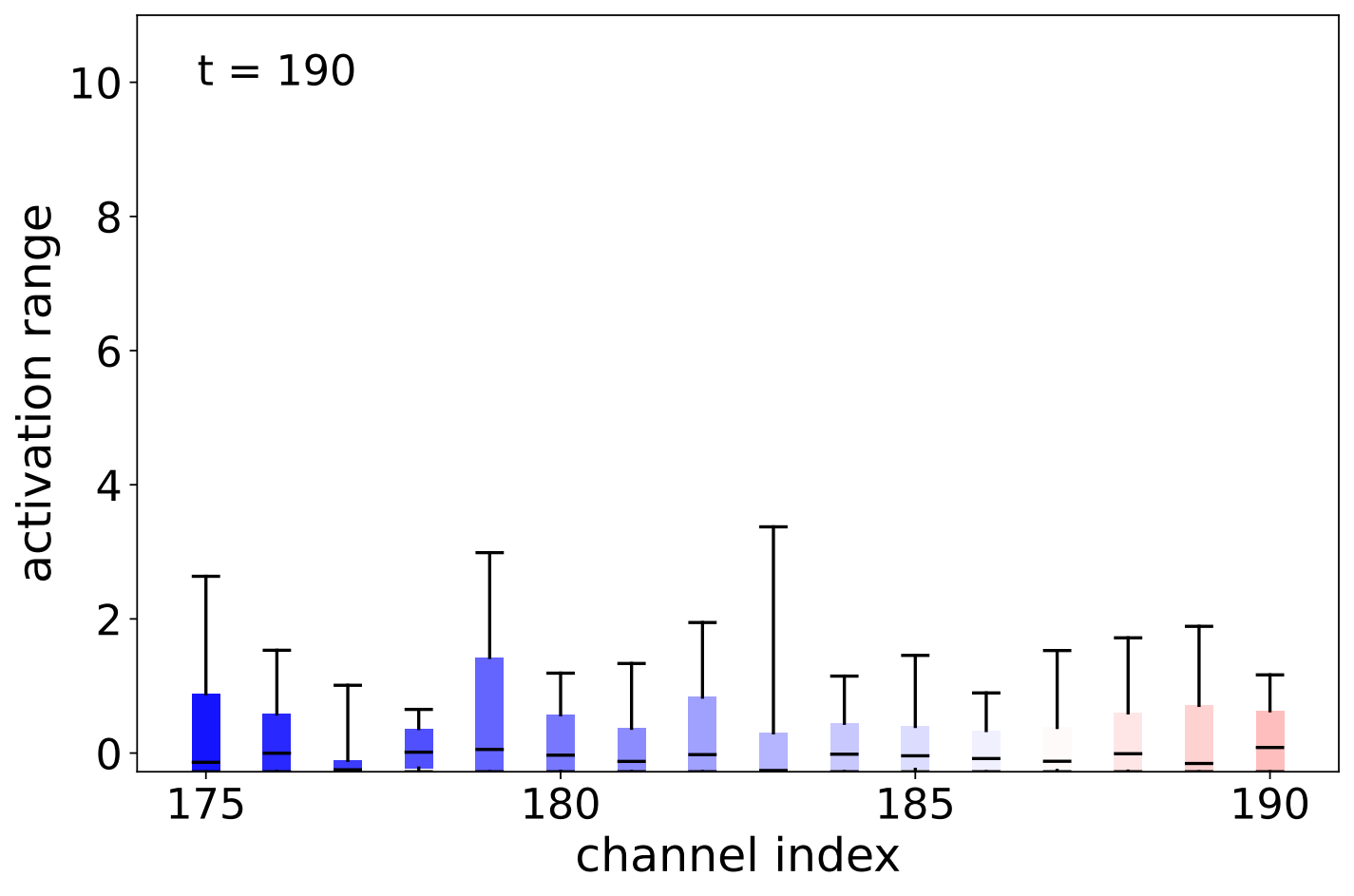}
            \label{fig:dist_sub2}
        \end{subfigure}
        \begin{subfigure}{.3\columnwidth}
            \centering
            \includegraphics[width=1\linewidth]{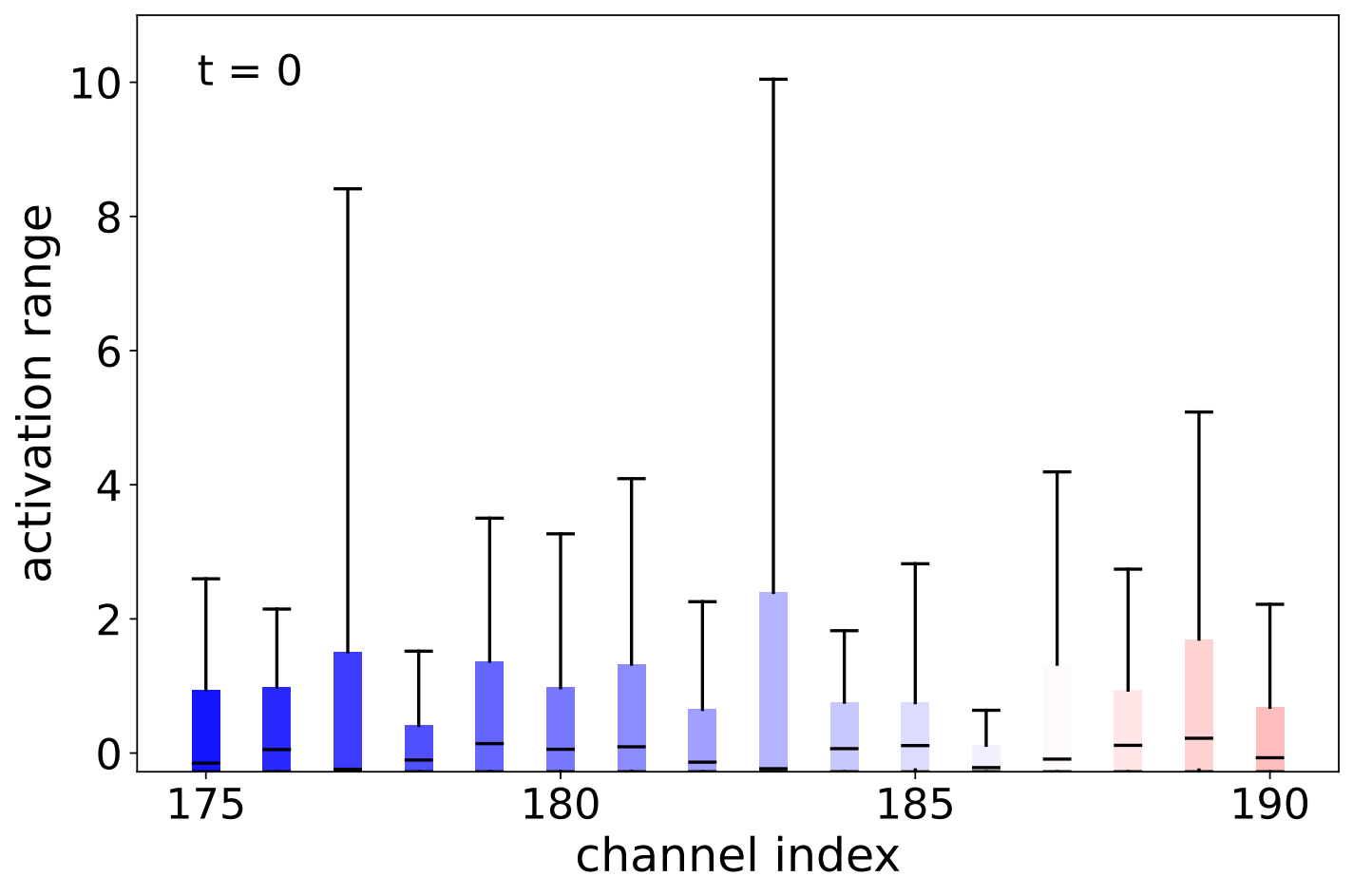}
            \label{fig:dist_sub3}
        \end{subfigure}
        \caption{Activation before reparameterization}
    \end{subfigure}
    \begin{subfigure}{2\columnwidth}
        \centering
        \begin{subfigure}{.3\columnwidth}
            \centering
            \includegraphics[width=1\linewidth]{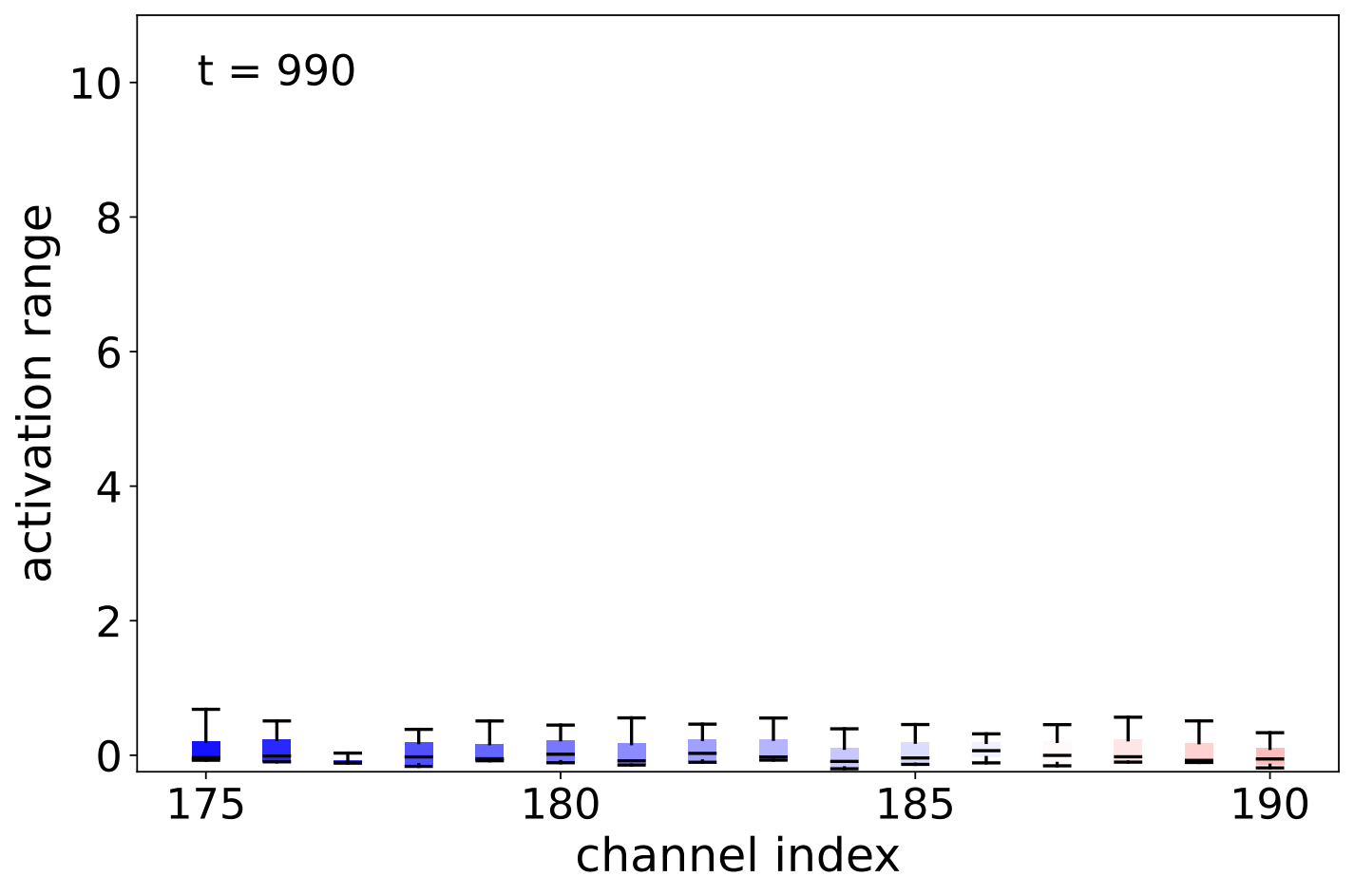}
            \label{fig:dist_sub4}
        \end{subfigure}
        \begin{subfigure}{.3\columnwidth}
            \centering
            \includegraphics[width=1\linewidth]{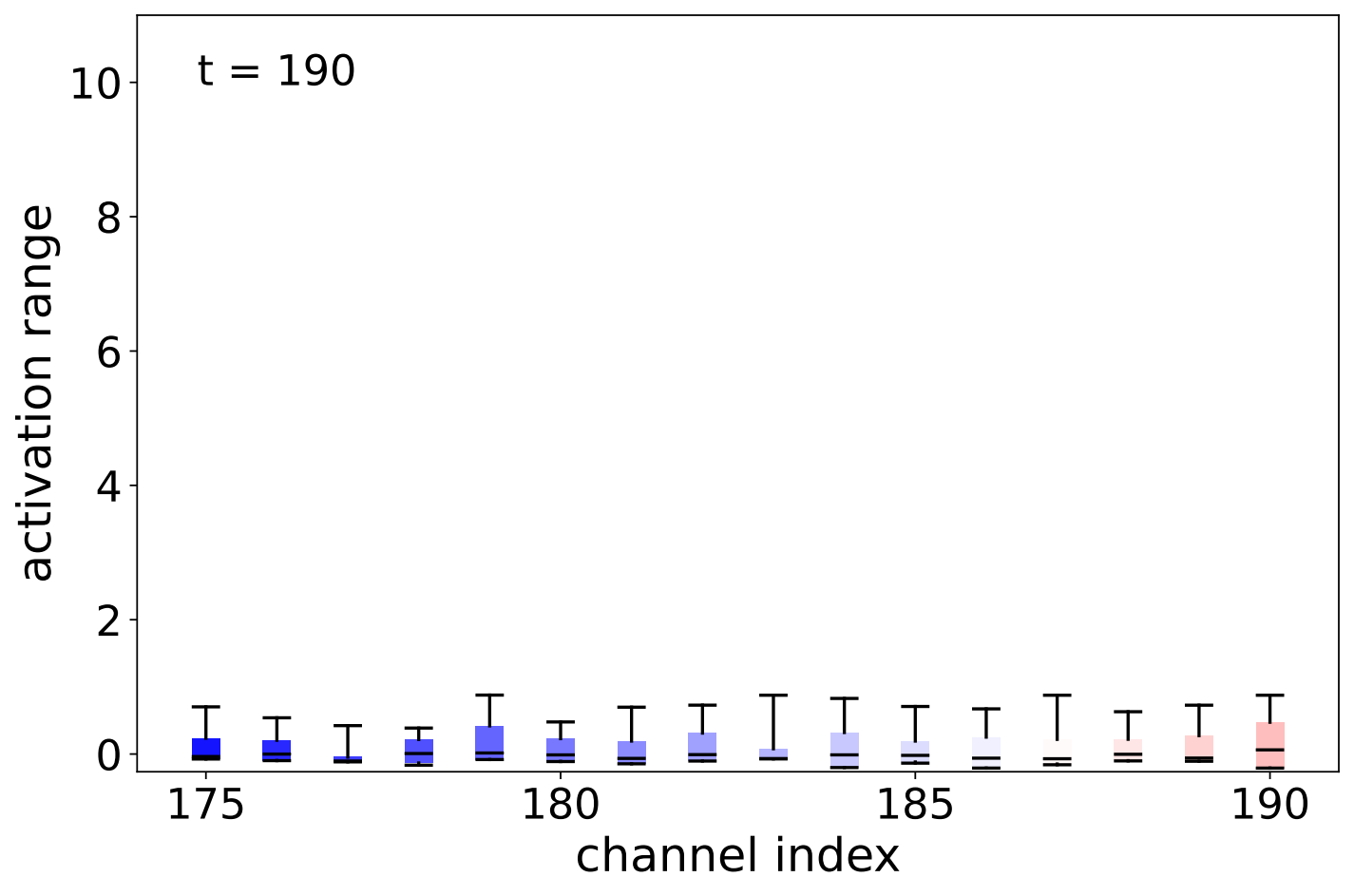}
            \label{fig:dist_sub5}
        \end{subfigure}
        \begin{subfigure}{.3\columnwidth}
            \centering
            \includegraphics[width=1\linewidth]{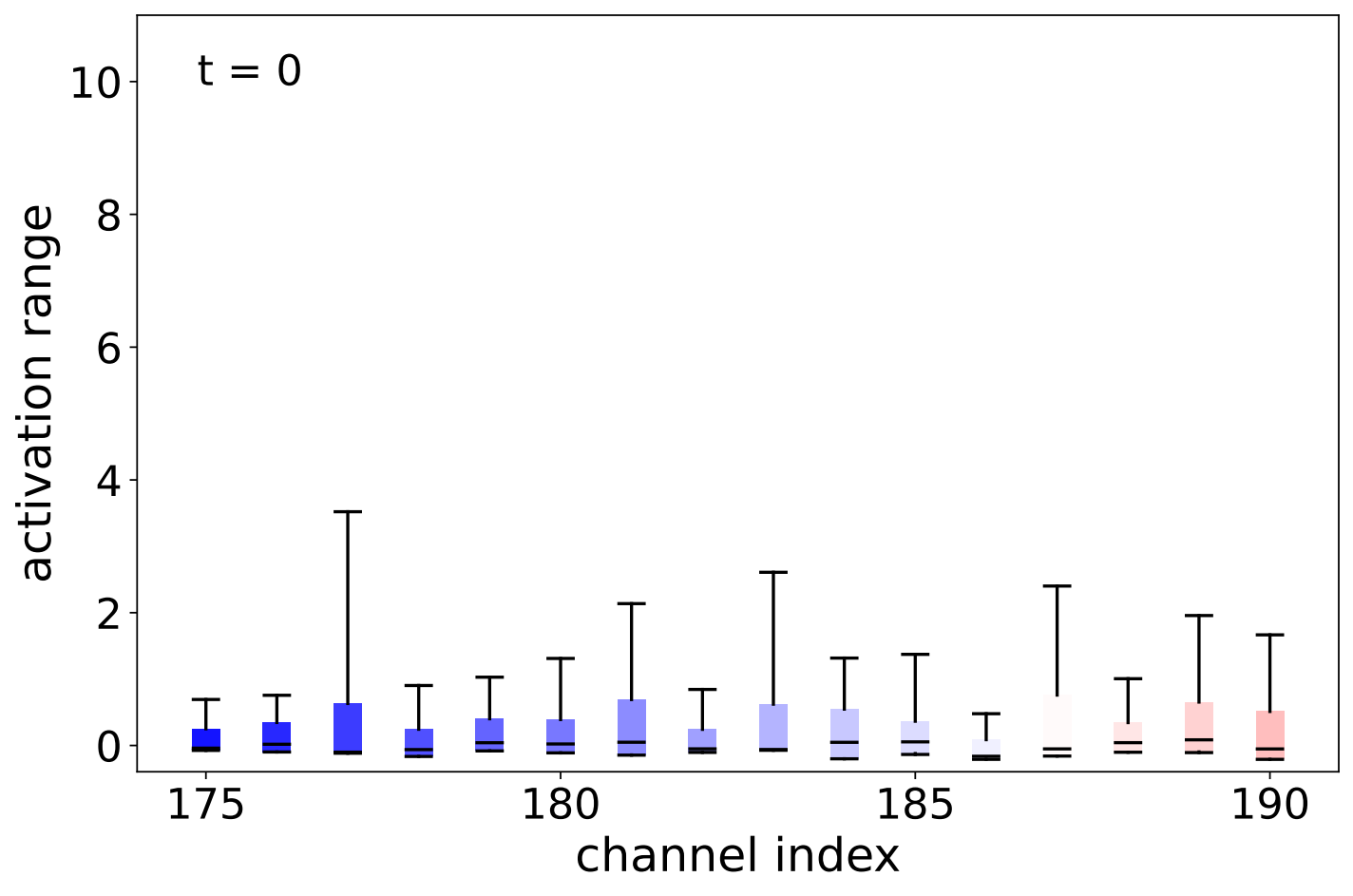}
            \label{fig:dist_sub6}
        \end{subfigure}
        \caption{Activation after reparameterization}
    \end{subfigure}
    \caption{Illustration of the effect of the reparameterization on the \textbf{up.0.block.0.conv1} layer  of the DDIM model. (a) shows the activation distribution of the original model, which suffers from a severe channel variance and outliers. By comparing (a) and (b), we can observe that the fluctuated range between channels is mitigated and the value range is 
    significantly narrowed, by adopting the proposed TCR module.}
    \label{fig:dist}
\end{figure*}

As shown in Fig. \ref{fig:dist}, the activation range is effectively restrained and balanced by our proposed TCR module, significantly facilitating the quantization process.

In practice, we observe that the performance of the proposed TCR module deteriorates when weights are quantized at low bit-widths, such as in the W4A8 setting. This degradation is primarily due to the greater sensitivity of weights to changes in value range compared to activations, while the enlarged weights are prone to incurring more pronounced errors. To address this issue, we define a clamping range $r_c$ to restrict the scaling vector to a narrow range. As displayed in Table \ref{tab:TCR-clamp-results}, a small clamp range will benefit the final performance of the TCR module.

\begin{algorithm}[!t]
    \caption{Progressively Aligned Reconstruction}
    \begin{algorithmic}[1]
        \Require The full-precision model $m_{f}$, the progressive rounds $p$, the number of reconstruction iterations $i$, and the Diffusion Sampler $sampler$.
        \Ensure Quantized model $m_{q_p}$.
        \State $cali \gets sampler(m_{f})$.
        \State $m_{q_0} \gets Initialization(m_{f}, cali)$ and perform the basic Adaround operation.
        \State $n \gets 0$.
        \While{$ n < p $}
            \State $cali \gets sample(m_{q_n})$.
            \State Optimize the quantization parameters $m_{q_{n+1}}$ by using Adaround based on the calibration set $cali$ with $i$ iterations. 
            \State $n \gets n + 1$.
        \EndWhile
        \State \Return $m_{q_p}$.
    \end{algorithmic}
\end{algorithm}
\begin{table}[!t]
    \centering
    \begin{tabular}{c|c|c|c}
    \hline 
    Bit-width & Clamp Range & FID($\downarrow$) & sFID($\downarrow$) \\    
    \hline
    W4A8 & 100 & 4.82 & 8.91 \\
    W4A8 & 50  & 4.76 & 8.95 \\
    W4A8 & 10 & 4.25 & 9.01 \\
    W4A8 & 5 & \textbf{4.14} & \textbf{9.06} \\
    W4A8 & 3 & 4.28 & 9.03 \\
    \hline
    \end{tabular}
    \caption{Ablation results of different clamp ranges on TCR on the CIFAR-10 dataset.}
    \label{tab:TCR-clamp-results}
\end{table}

\subsection{Additional Details and Results of PAR}
\begin{figure*}[!t]
    \centering
    \begin{subfigure}{.66\columnwidth}
        \centering
        \includegraphics[width=\columnwidth]{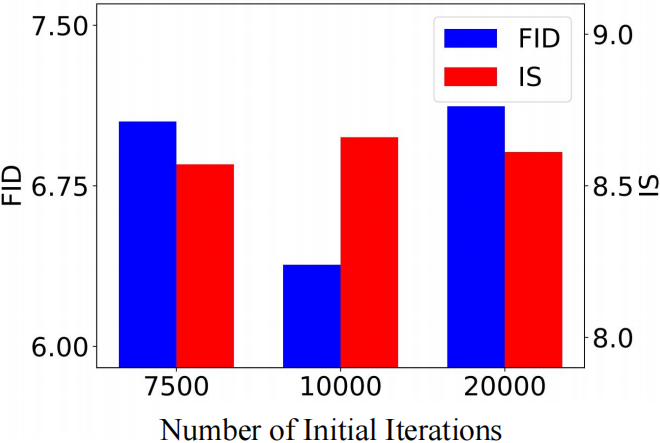}
        \caption{}
        \label{fig:par_sub1}
    \end{subfigure}
    \begin{subfigure}{.66\columnwidth}
        \centering
        \includegraphics[width=\columnwidth]{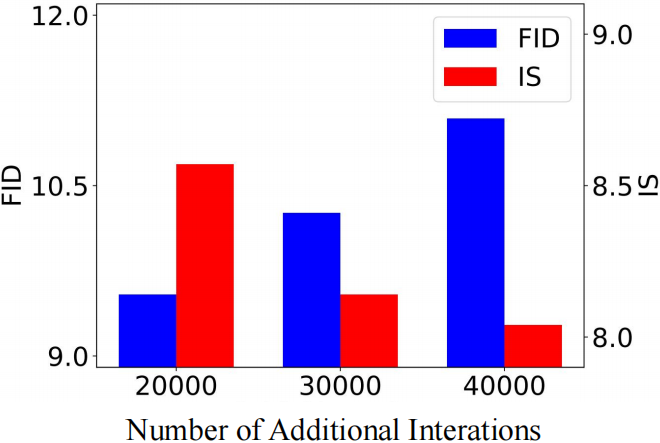}
        \caption{}
        \label{fig:par_sub2}
    \end{subfigure}
    \begin{subfigure}{.66\columnwidth}
        \centering
        \includegraphics[width=\columnwidth]{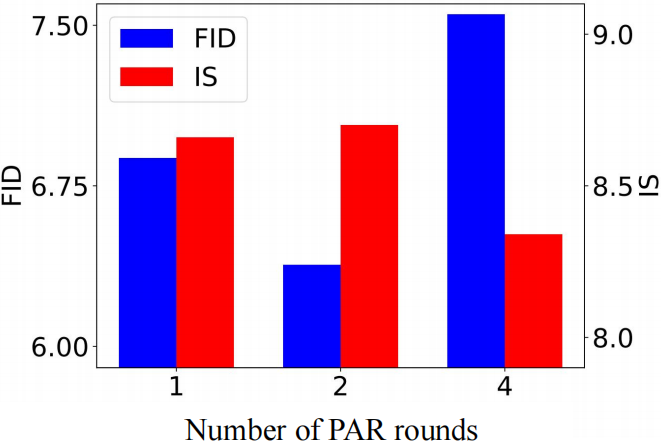}
        \caption{}
        \label{fig:par_sub3}
    \end{subfigure}

    \caption{Ablation results on PAR w.r.t. (a) the number of initial iterations, (b) the number of additional reconstruction iterations, and (c) the number of PAR rounds. }
    \label{fig:par}
\end{figure*}

The overall pipeline of the proposed progressively aligned reconstruction (PAR) is summarized in Algorithm 1. Moreover, as shown in Fig.~\ref{fig:par}(a), the basic Adaround methods yield inferior performance. However, resampling the calibration set improves the  reconstruction. We also provide additional experimental results on the hyper-parameters of PAR, including the additional number of iterations and experimental rounds. As shown in Fig. \ref{fig:par_sub1} (b) and (c), the proposed PAR achieves the best performance with 10,000 iterations and 2 additional PAR rounds.

\subsection{More Qualitative Results}
\begin{figure*}[!t]
    \centering
    \begin{subfigure}[ht]{.165\columnwidth}
        \centering
        \includegraphics[width=\linewidth]{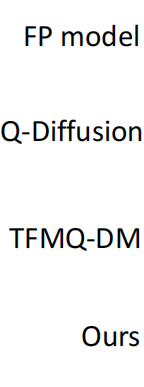}
        \label{fig:vis_title}
    \end{subfigure}
    \begin{subfigure}[ht]{.9\columnwidth}
        \centering
        \includegraphics[width=\linewidth]{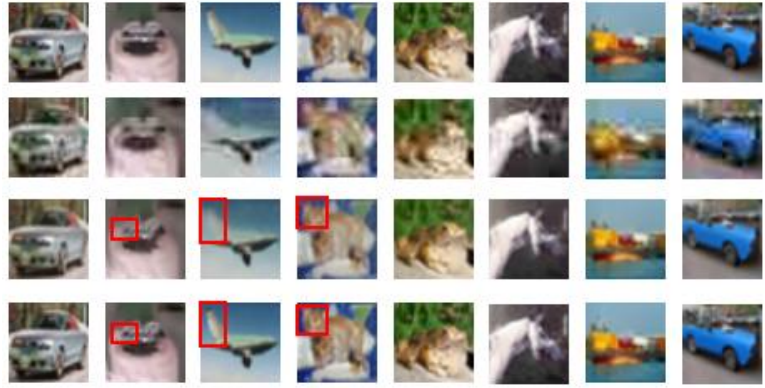}
        \caption{W6A6}
        \label{fig:vis_w6a6}
    \end{subfigure}
    \hfill
    \begin{subfigure}[ht]{.9\columnwidth}
        \centering
        \includegraphics[width=1\linewidth]{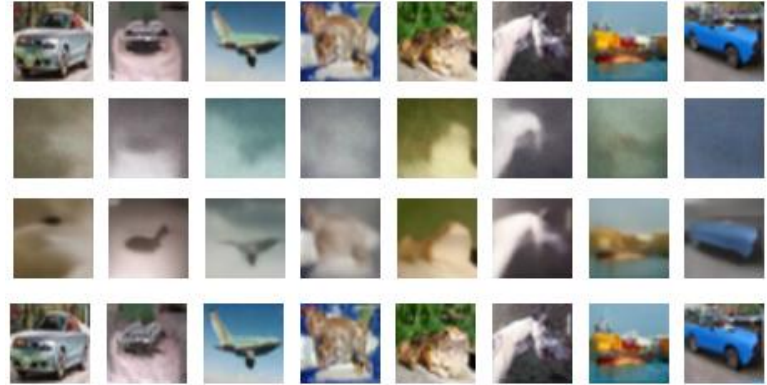}
        \caption{W4A4}
        \label{fig:vis_w4a4}
    \end{subfigure}
    \caption{Qualitative results on CIFAR-10 based on DDIM. The resolution of images is 32x32.}
    \label{fig:vis_all}
\end{figure*}

\begin{figure*}[!t]
    \centering
    \begin{subfigure}{2\columnwidth}
        \centering
        \includegraphics[width=.2\linewidth]{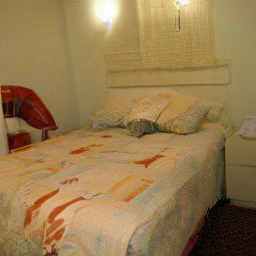}
        \includegraphics[width=.2\linewidth]{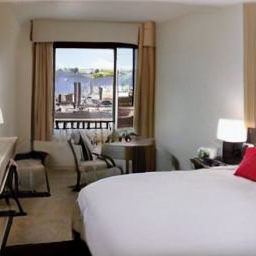}
        \includegraphics[width=.2\linewidth]{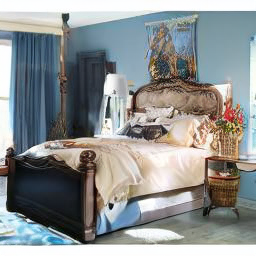}
        \includegraphics[width=.2\linewidth]{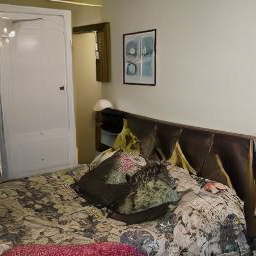}
        \caption{Full precision model}
    \end{subfigure}
    \begin{subfigure}{2\columnwidth}
        \centering
        \includegraphics[width=.2\linewidth]{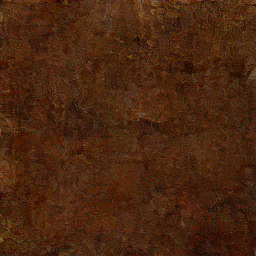}
        \includegraphics[width=.2\linewidth]{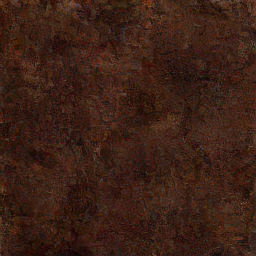}
        \includegraphics[width=.2\linewidth]{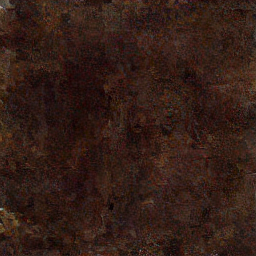}
        \includegraphics[width=.2\linewidth]{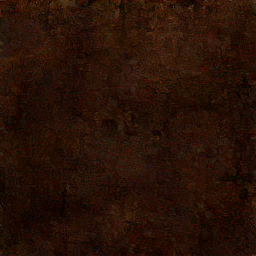}
        \caption{Q-Diffusion (W4A4)}
    \end{subfigure}
    \begin{subfigure}{2\columnwidth}
        \centering
        \includegraphics[width=.2\linewidth]{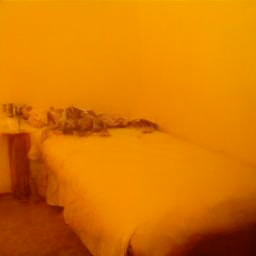}
        \includegraphics[width=.2\linewidth]{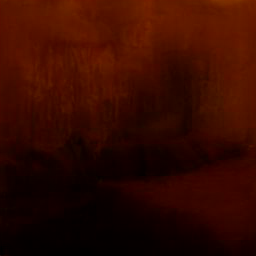}
        \includegraphics[width=.2\linewidth]{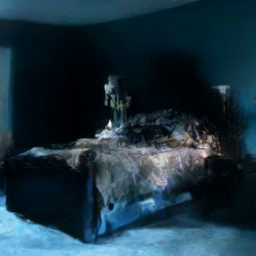}
        \includegraphics[width=.2\linewidth]{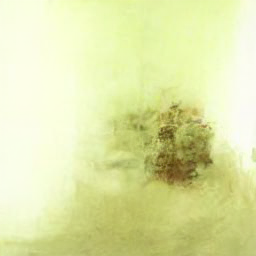}
        \caption{TFMQ-DM (W4A4)}
    \end{subfigure}
    \begin{subfigure}{2\columnwidth}
        \centering
        \includegraphics[width=.2\linewidth]{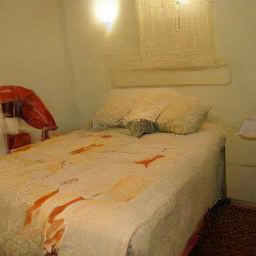}
        \includegraphics[width=.2\linewidth]{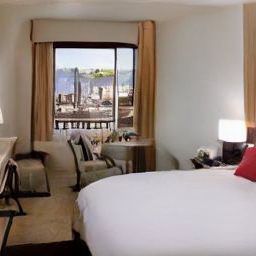}
        \includegraphics[width=.2\linewidth]{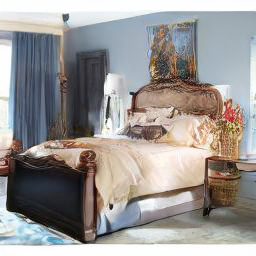}
        \includegraphics[width=.2\linewidth]{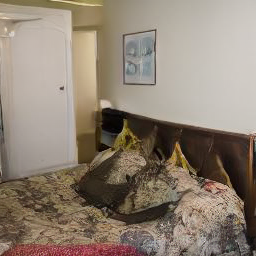}
        \caption{\textbf{Ours} (W4A4)}
    \end{subfigure}
    \caption{Qualitative results on LSUN-Bedrooms based on LDM-4. The resolution of images is 256x256.}
    \label{fig:bed_w4a4_vis}
\end{figure*}

\begin{figure*}[!t]
    \centering
    \begin{subfigure}{2\columnwidth}
        \centering
        \includegraphics[width=.2\linewidth]{fig/vis/bedrooms/fp/fp_0.png}
        \includegraphics[width=.2\linewidth]{fig/vis/bedrooms/fp/fp_1.png}
        \includegraphics[width=.2\linewidth]{fig/vis/bedrooms/fp/fp_2.png}
        \includegraphics[width=.2\linewidth]{fig/vis/bedrooms/fp/fp_3.png}
        \caption{Full precision model}
    \end{subfigure}
    \begin{subfigure}{2\columnwidth}
        \centering
        \includegraphics[width=.2\linewidth]{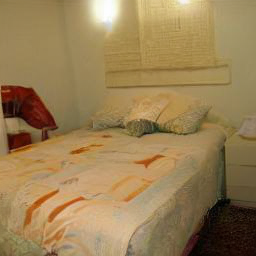}
        \includegraphics[width=.2\linewidth]{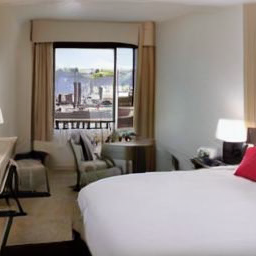}
        \includegraphics[width=.2\linewidth]{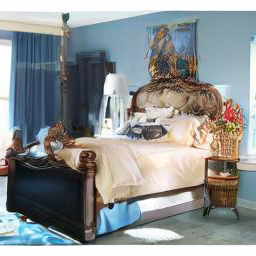}
        \includegraphics[width=.2\linewidth]{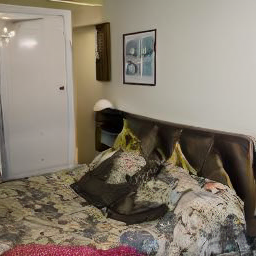}
        \caption{Q-Diffusion (W4A8)}
    \end{subfigure}
    \begin{subfigure}{2\columnwidth}
        \centering
        \includegraphics[width=.2\linewidth]{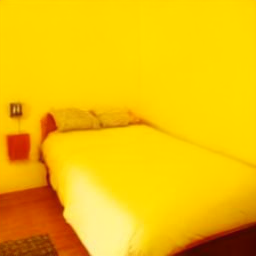}
        \includegraphics[width=.2\linewidth]{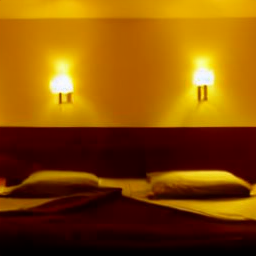}
        \includegraphics[width=.2\linewidth]{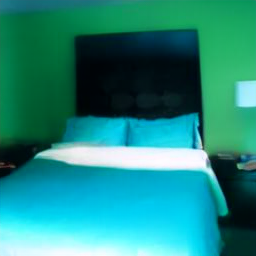}
        \includegraphics[width=.2\linewidth]{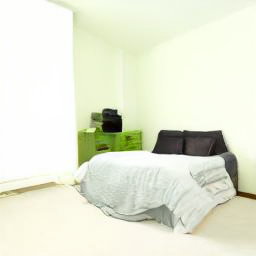}
        \caption{TFMQ-DM (W4A8)}
    \end{subfigure}
    \begin{subfigure}{2\columnwidth}
        \centering
        \includegraphics[width=.2\linewidth]{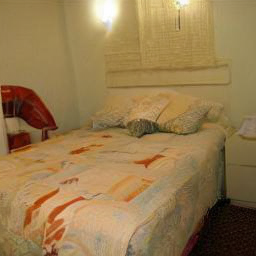}
        \includegraphics[width=.2\linewidth]{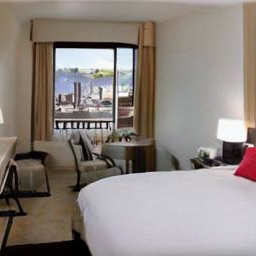}
        \includegraphics[width=.2\linewidth]{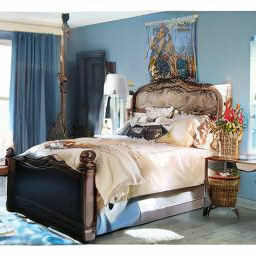}
        \includegraphics[width=.2\linewidth]{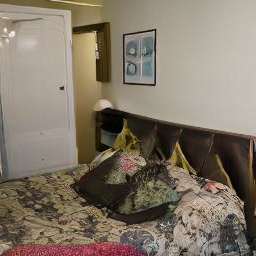}
        \caption{Ours (W4A8)}
    \end{subfigure}
    \caption{Qualitative results on LSUN-Bedrooms based on LDM-4. The resolution of images is 256x256.}
    \label{fig:bed_w4a8_vis}
\end{figure*}

\begin{figure*}[!t]
    \centering
    \begin{subfigure}{2\columnwidth}
        \centering
        \includegraphics[width=.2\linewidth]{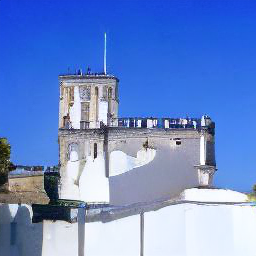}
        \includegraphics[width=.2\linewidth]{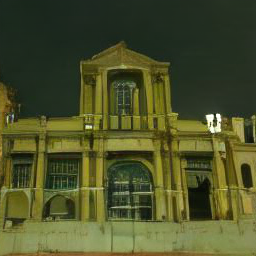}
        \includegraphics[width=.2\linewidth]{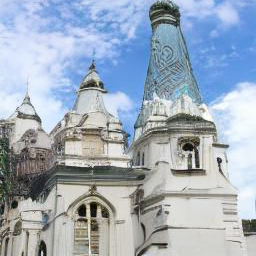}
        \includegraphics[width=.2\linewidth]{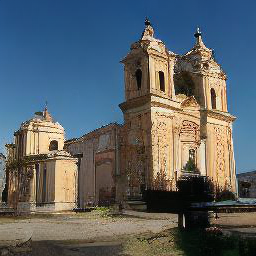}
        \caption{Full precision model}
    \end{subfigure}
    \begin{subfigure}{2\columnwidth}
        \centering
        \includegraphics[width=.2\linewidth]{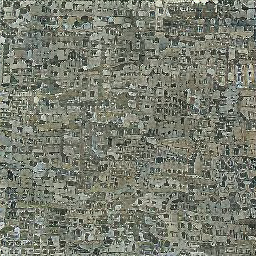}
        \includegraphics[width=.2\linewidth]{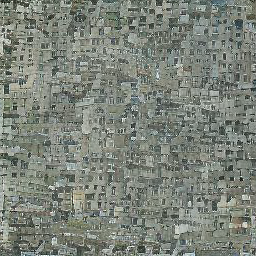}
        \includegraphics[width=.2\linewidth]{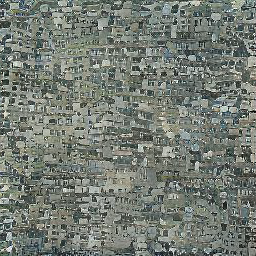}
        \includegraphics[width=.2\linewidth]{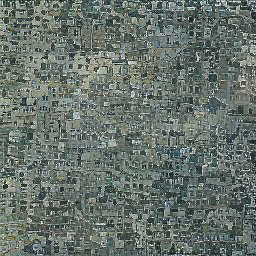}
        \caption{Q-Diffusion (W4A4)}
    \end{subfigure}
    \begin{subfigure}{2\columnwidth}
        \centering
        \includegraphics[width=.2\linewidth]{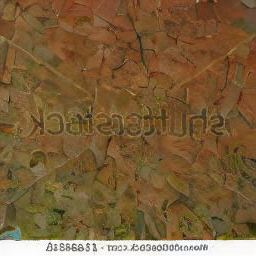}
        \includegraphics[width=.2\linewidth]{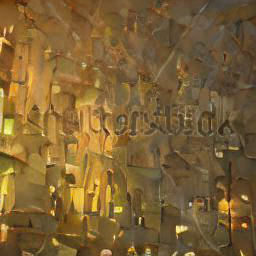}
        \includegraphics[width=.2\linewidth]{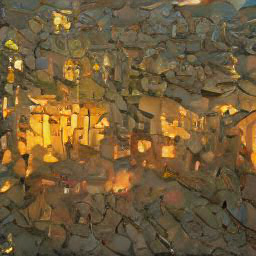}
        \includegraphics[width=.2\linewidth]{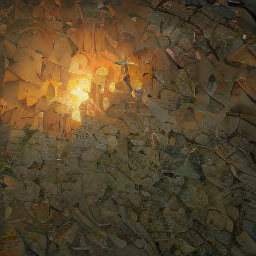}
        \caption{TFMQ-DM (W4A4)}
    \end{subfigure}
    \begin{subfigure}{2\columnwidth}
        \centering
        \includegraphics[width=.2\linewidth]{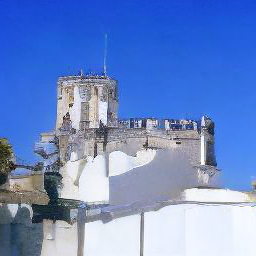}
        \includegraphics[width=.2\linewidth]{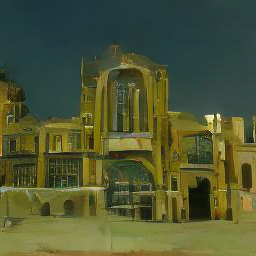}
        \includegraphics[width=.2\linewidth]{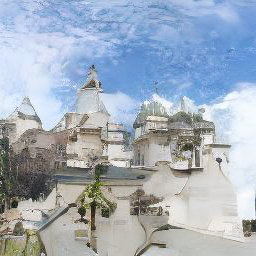}
        \includegraphics[width=.2\linewidth]{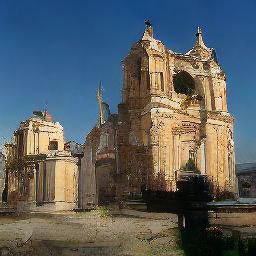}
        \caption{\textbf{Ours} (W4A4)}
    \end{subfigure}
    \caption{Qualitative results on LSUN-Churches based on LDM-4. The resolution of images is 256x256.}
    \label{fig:churches_w4a4_vis}
\end{figure*}

\begin{figure*}[ht]
    \centering
    \begin{subfigure}{2\columnwidth}
        \centering
        \includegraphics[width=.2\linewidth]{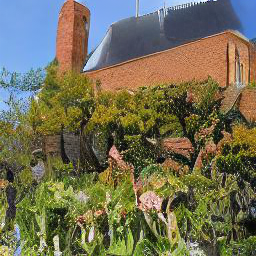}
        \includegraphics[width=.2\linewidth]{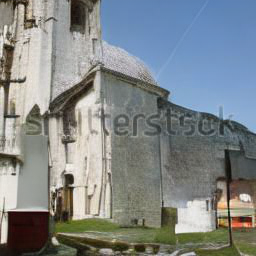}
        \includegraphics[width=.2\linewidth]{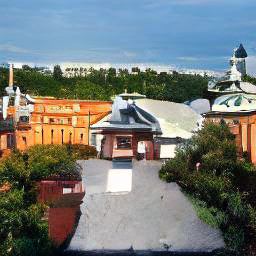}
        \includegraphics[width=.2\linewidth]{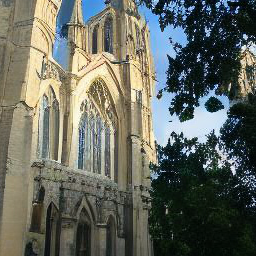}
        \caption{Full precision model}
    \end{subfigure}
    \begin{subfigure}{2\columnwidth}
        \centering
        \includegraphics[width=.2\linewidth]{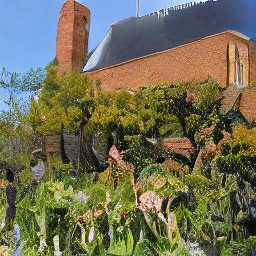}
        \includegraphics[width=.2\linewidth]{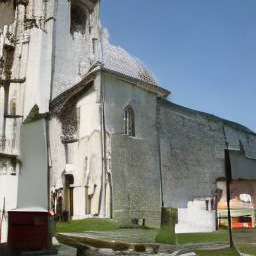}
        \includegraphics[width=.2\linewidth]{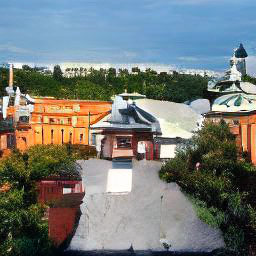}
        \includegraphics[width=.2\linewidth]{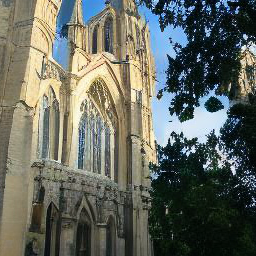}
        \caption{Q-Diffusion (W4A8)}
    \end{subfigure}
    \begin{subfigure}{2\columnwidth}
        \centering
        \includegraphics[width=.2\linewidth]{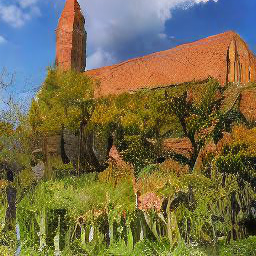}
        \includegraphics[width=.2\linewidth]{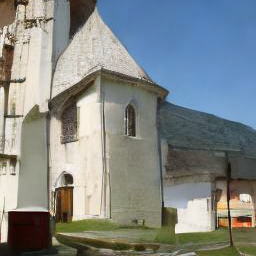}
        \includegraphics[width=.2\linewidth]{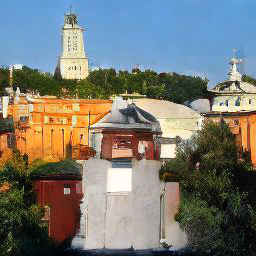}
        \includegraphics[width=.2\linewidth]{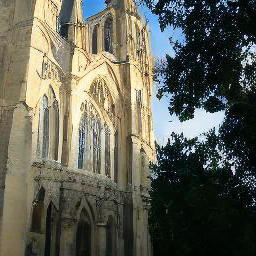}
        \caption{TFMQ-DM (W4A8)}
    \end{subfigure}
    \begin{subfigure}{2\columnwidth}
        \centering
        \includegraphics[width=.2\linewidth]{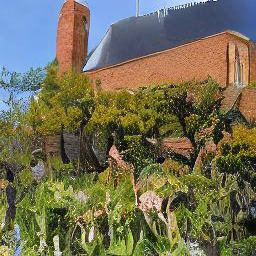}
        \includegraphics[width=.2\linewidth]{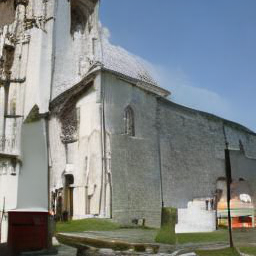}
        \includegraphics[width=.2\linewidth]{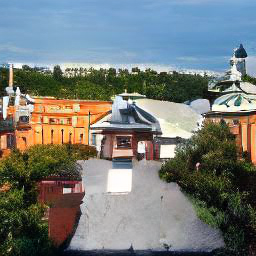}
        \includegraphics[width=.2\linewidth]{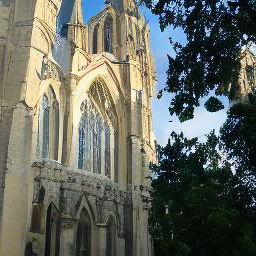}
        \caption{\textbf{Ours} (W4A8)}
    \end{subfigure}
    \caption{Qualitative results on LSUN-Churches based on LDM-4. The resolution of images is 256x256.}
    \label{fig:church_w4a8_vis}
\end{figure*}

\begin{figure*}[ht]
    \centering
    \begin{subfigure}{2\columnwidth}
        \centering
        \includegraphics[width=.13\linewidth]{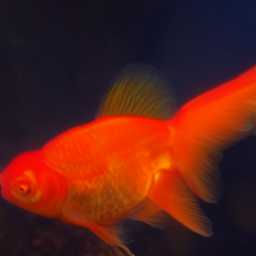}
        \includegraphics[width=.13\linewidth]{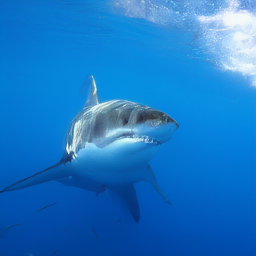}
        \includegraphics[width=.13\linewidth]{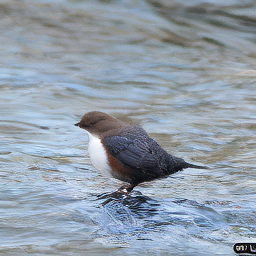}
        \includegraphics[width=.13\linewidth]{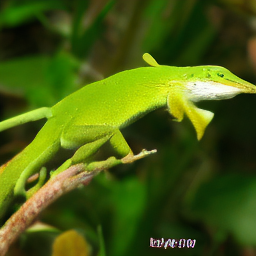}
        \includegraphics[width=.13\linewidth]{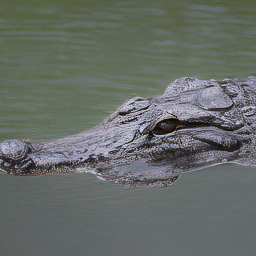}
        \includegraphics[width=.13\linewidth]{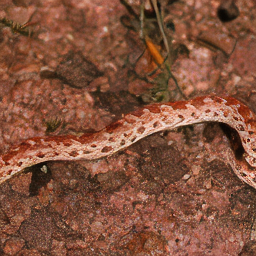}
        \caption{Full precision model}
    \end{subfigure}
    \begin{subfigure}{2\columnwidth}
        \centering
        \includegraphics[width=.13\linewidth]{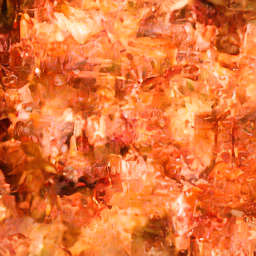}
        \includegraphics[width=.13\linewidth]{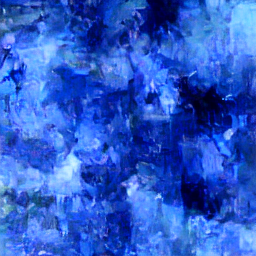}
        \includegraphics[width=.13\linewidth]{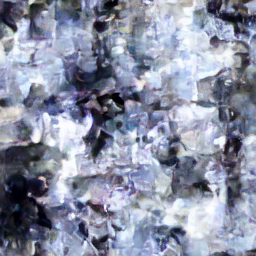}
        \includegraphics[width=.13\linewidth]{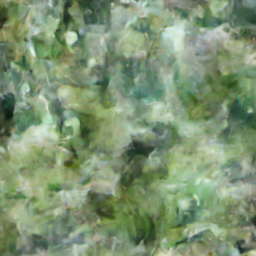}
        \includegraphics[width=.13\linewidth]{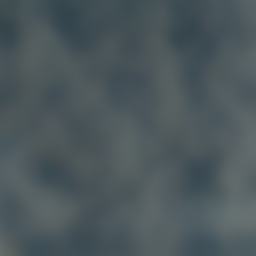}
        \includegraphics[width=.13\linewidth]{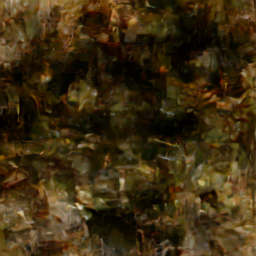}
        \caption{Q-Diffusion (W4A4)}
    \end{subfigure}
    \begin{subfigure}{2\columnwidth}
        \centering
        \includegraphics[width=.13\linewidth]{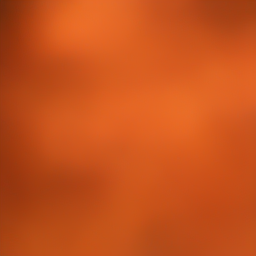}
        \includegraphics[width=.13\linewidth]{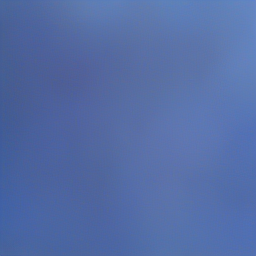}
        \includegraphics[width=.13\linewidth]{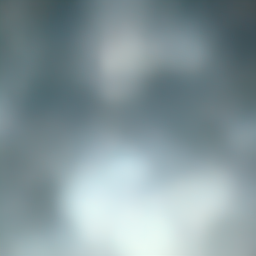}
        \includegraphics[width=.13\linewidth]{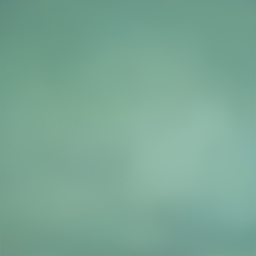}
        \includegraphics[width=.13\linewidth]{fig/vis/imagenet/W4A4/TFMQ_500.png}
        \includegraphics[width=.13\linewidth]{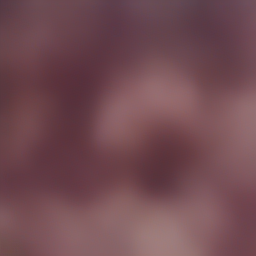}
        \caption{TFMQ-DM (W4A4)}
    \end{subfigure}
    \begin{subfigure}{2\columnwidth}
        \centering
        \includegraphics[width=.13\linewidth]{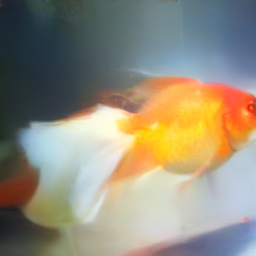}
        \includegraphics[width=.13\linewidth]{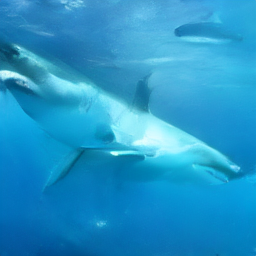}
        \includegraphics[width=.13\linewidth]{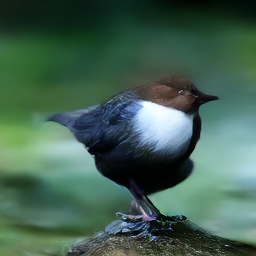}
        \includegraphics[width=.13\linewidth]{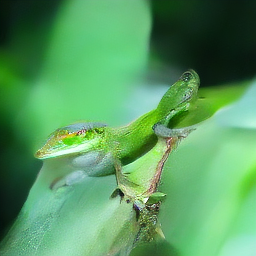}
        \includegraphics[width=.13\linewidth]{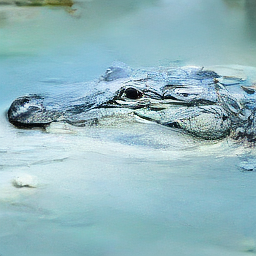}
        \includegraphics[width=.13\linewidth]{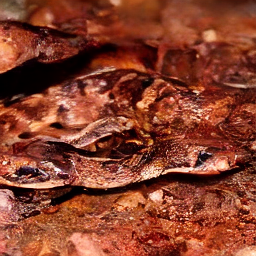}
        \caption{\textbf{Ours} (W4A4)}
    \end{subfigure}
    \caption{Qualitative results on ImageNet based on LDM-4. The resolution of images is 256x256.}
    \label{fig:imagenet_w4a4_vis}
\end{figure*}

In addition to Fig.~3 of the main body, we provide more qualitative results on different datasets in Fig. \ref{fig:vis_all} to Fig. \ref{fig:imagenet_w4a4_vis}. The experiment settings are the same as the used in the experimental parts of the main body. We re-implement Q-Diffusion and TFMQ-DM for comparisons. As illustrated in Fig. \ref{fig:vis_all} to Fig. \ref{fig:imagenet_w4a4_vis}, our method generates more realistic images in most cases, showing its superiority.

